\documentclass[sn-nature]{sn-jnl}%

\usepackage{graphicx}%
\usepackage{multirow}%
\usepackage{amsmath,amssymb,amsfonts}%
\usepackage{amsthm}%
\usepackage{mathrsfs}%
\usepackage[title]{appendix}%
\usepackage{xcolor}%
\usepackage{textcomp}%
\usepackage{manyfoot}%
\usepackage{booktabs}%
\usepackage{algorithm}%
\usepackage{algorithmicx}%
\usepackage{algpseudocode}%
\usepackage{listings}%
\usepackage{fontawesome5}
\usepackage{makecell}
\usepackage{caption}
\usepackage{hyperref}
\usepackage{wrapfig}
\usepackage{subcaption}
\usepackage{float}
\usepackage{doi}

\newcommand{\notsure}[1]{\textcolor{black}{#1}}
\newcommand{\revise}[1]{\textcolor{black}{#1}}
\newcommand{\final}[1]{\textcolor{black}{#1}}
\newcommand{\editorial}[1]{\textcolor{black}{#1}}

\begin{document}

\title[Article Title]{A Large Model for Non-invasive and Personalized Management of Breast Cancer from Multiparametric MRI}

\author[1,2]{\fnm{Luyang} \sur{Luo}}

\author[3]{\fnm{Mingxiang} \sur{Wu}}

\author[4]{\fnm{Mei} \sur{Li}}

\author[1]{\fnm{Yi} \sur{Xin}}

\author[5]{\fnm{Qiong} \sur{Wang}}
 
\author[6]{\fnm{Varut} \sur{Vardhanabhuti}}

\author[7]{\fnm{Winnie CW} \sur{Chu}}

\author*[8]{\fnm{Zhenhui} \sur{Li}}\email{lizhenhui@kmmu.edu.cn}

\author*[9,10]{\fnm{Juan} \sur{Zhou}}\email{ zjuan122@163.com}

\author[2]{\fnm{Pranav} \sur{Rajpurkar}}

\author*[1,11,12,13]{\fnm{Hao} \sur{Chen}}\email{jhc@cse.ust.hk}

\affil[1]{\orgdiv{Department of Computer Science and Technology}, \orgname{The Hong Kong University of Science and Technology}, \city{Hong Kong}, \country{China}}

\affil[2]{\orgdiv{Department of Biomedical Informatics}, \orgname{Harvard University}, \city{Boston}, \country{U.S.A}}

\affil[3]{\orgdiv{Department of Radiology}, \orgname{Shenzhen People's Hospital}, \city{Shenzhen}, \country{China}}

\affil[4]{\orgdiv{Department of Radiology}, \orgname{PLA Middle Military Command General Hospital}, \city{Wuhan}, \country{China}}

\affil[5]{\orgdiv{Shenzhen Institute of Advanced Technology}, \orgname{Chinese Academy of Sciences}, \city{Shenzhen}, \country{China}}

\affil[6]{\orgdiv{Department of Diagnostic Radiology}, \orgname{Li Ka Shing Faculty of Medicine, The University of Hong Kong}, \city{Hong Kong}, \country{China}}

\affil[7]{\orgdiv{Department of Imaging and Interventional Radiology}, \orgname{The Chinese University of Hong Kong}, \city{Hong Kong}, \country{China}}

\affil[8]{\orgdiv{Department of Radiology},\orgname{the Third Affiliated Hospital of Kunming Medical
University, Yunnan Cancer Hospital, Yunnan Cancer Center}, \city{Kunming}, \country{China}}

\affil[9]{\orgdiv{Department of Radiology}, \orgname{5th Medical Center of Chinese PLA General Hospital}, \city{Beijing}, \country{China}}

\affil[10]{\orgdiv{The Second School of Clinical Medicine}, \orgname{Southern Medical University}, \city{Guangzhou}, \country{China}}

\affil[11]{\orgdiv{Department of Chemical and Biological Engineering}, \orgname{The Hong Kong University of Science and Technology}, \city{Hong Kong}, \country{China}}

\affil[12]{\orgdiv{Division of Life Science}, \orgname{The Hong Kong University of Science and Technology}, \city{Hong Kong}, \country{China}}

\affil[13]{\orgdiv{State Key Laboratory of Molecular Neuroscience}, \orgname{The Hong Kong University of Science and Technology}, \city{Hong Kong}, \country{China}}

\abstract{
Breast \editorial{Magnetic Resonance Imaging} (MRI) demonstrates the highest sensitivity for breast cancer detection among imaging modalities and is standard practice for high-risk women. Interpreting the multi-sequence MRI is time-consuming and prone to subjective variation. We develop a large mixture-of-modality-experts model (MOME) that integrates multiparametric MRI information within a unified structure, leveraging breast MRI scans from 5,205 female patients in China for model development and validation. MOME matches four senior radiologists' performance in identifying breast cancer and outperforms a junior radiologist. \editorial{The model is able to reduce unnecessary biopsies in Breast Imaging-Reporting and Data System} (BI-RADS) 4 patients, classify triple-negative breast cancer, and predict pathological complete response to neoadjuvant chemotherapy. MOME further supports inference with missing modalities and provides decision explanations by highlighting lesions and measuring modality contributions. To summarize, MOME exemplifies an accurate and robust multimodal model for noninvasive, personalized management of breast cancer patients via multiparametric MRI.
}


\maketitle
\section*{Introduction}\label{sec:intro}
Breast cancer is the primary cause of cancer mortality in females worldwide \cite{bray2022global}. Early detection and accurate, systematic treatment are crucial for reducing mortality \cite{HARBECK20171134}.
\revise{Personalization of breast cancer treatment can lead to improved patient outcomes, which can be benefited by molecular subtyping and treatment response prediction for treatment selection guidance \cite{goldhirsch2013personalizing,zhang2023predicting}. 
For example, ‘triple-negative’ breast cancer patients showed improved event-free survival with pathologic complete response (pCR) after neoadjuvant chemotherapy (NACT) \cite{spring2020pathologic}.}
%requiring precise malignancy screening and guidance by molecular subtyping and treatment response estimation , informed by breast cancer examinations and analytics. 
Breast magnetic resonance imaging (MRI) is the radiology technique with the highest sensitivity for breast cancer detection and plays an indispensable role in breast cancer screening and staging for high-risk women \cite{mann2019breast}, holding promise as a non-invasive investigation approach.
It is also recommended as a screening technique for women with dense breasts, a condition that is prevalent among women in Eastern populations, such as those in China \cite{sung2018breast}.
Currently, reading breast MRI is majorly based on the American College of Radiology Breast Imaging Reporting and Data System (BI-RADS) \cite{american2013birads}, which requires comprehension of the information from multiparametric MRI data, routinely including the T1-weighted dynamic contrast-enhanced sequence (DCE-MRI), the T2-weighted imaging (T2WI), and the diffusion-weighted imaging (DWI) to improve differentiation of breast lesions.

\revise{Artificial Intelligence (AI), typically deep learning \cite{lecun2015deep}, has shown remarkable progress in healthcare \cite{rajpurkar2022ai}, including breast cancer imaging \cite{luo2024deep}, where imaging features are largely used to generate quantitative biomarkers for breast cancer analytic \cite{zhang2023radiomics}.}
Despite the increasing diagnosis accuracy reported for AI-based breast cancer diagnosis from MRI, existing studies were mostly based on DCE-MRI \cite{zhou2019weakly,jiang2021artificial,witowski2022improving,verburg2022deep}, one single modality that often leads to high sensitivity with moderate specificity.
However, the diagnosis and prognosis based on breast MRI is routinely a multiparametric process, and how other sequences may help in increasing the specificity and overall accuracy remains to be explore \cite{monticciolo2018breast,peters2008meta,youn2024diagnostic,bluemke2004magnetic}.
Apart from the challenges brought by the collection of the multi-sequence data, multimodal integration also faces technical difficulties raised by the heterogeneity and high dimensionality \cite{acosta2022multimodal}.
Specifically, different modalities often require tailored representation learners, and a meticulously designed fusion module is demanded for heterogeneous information interaction modeling.
Such an architecture is absent in the literature on AI-based breast MRI analysis, and its clinical value remains to be further investigated.

As a recent generation of AI, foundation models (FMs) \cite{bommasani2021opportunities} were heralded as a promising solution to comprehend the heterogeneous multimodal information \cite{moor2023foundation}.
Particularly, FMs are developed with massive, diverse datasets and enable generalizability on multifaceted tasks by the large-scale pre-training paradigm \cite{radford2021learning,wang2023image}.
Moreover, the ability of FMs in unifying multimodal representations can be facilitated with a unified Transformer structure \cite{vaswani2017attention} by extensively modeling the cross-range dependencies among the input tokens. 
In medical image analysis, recent studies also demonstrated that these methods could match medical specialists' performance, such as chest X-rays \cite{zhou2022generalized,tiu2022expert}, pathology images \cite{chen2024towards,lu2024visual,xu2024whole}, and transcriptomics \cite{hao2024large}.
Nevertheless, FMs often contain billions of parameters that need to be pre-trained from million-scale datasets, which is impractical for situations with less data.
Under such circumstances, parameter-efficient fine-tuning provides the feasibility of leveraging the pre-training knowledge by adapting from existing FMs with a confined scale of trainable weights \cite{ding2023parameter}.
In spite of this, adapting the foundation model knowledge learned from 2D natural images for multiparametric MRI analysis encounters a significant domain gap, particularly with the increase in the number of modalities and dimensions.

\begin{figure*}[t]
    \centering
\includegraphics[width=\textwidth]{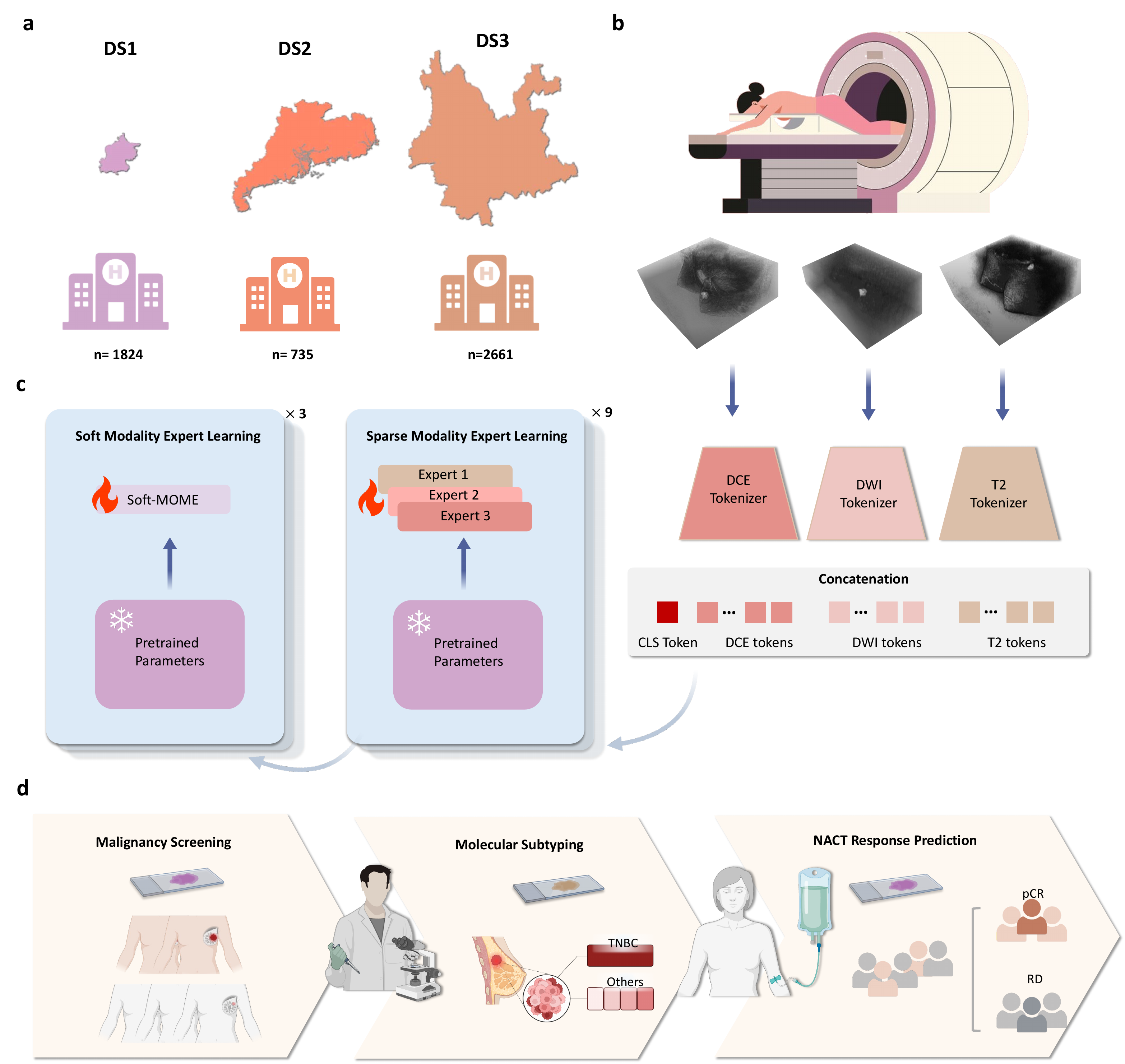}
    \caption{\textbf{Overview of the multiparametric breast MRI-based study design.} a. Data collection from three different hospitals, covering the population of the north, southeast, and southwest of China. b. The generation of multiparametric breast MRI, where T2-weighted MRI, Diffusion-weighted MRI, and DCE-MRI were mainly used in this study. c. MOME first takes multi-parametric MRI as input. Then, based on pre-trained foundation model parameters, mixture of sparse modality experts and soft modality experts are leveraged for unimodal feature extraction and multimodal information integration. d. MOME can be used for malignancy screening, molecular subtyping, and NACT response prediction, offering non-invasive personalized management for breast cancer patients. DS1 = dataset 1; DS2 = dataset 2; DS3 = dataset 3; MOME = mixture of modality experts; DCE = dynamic contrast-enhanced; DWI = diffusion-weighted Imaging; NACT = neoadjuvant chemotherapy. \editorial{Multiple elements were created with BioRender.com (\url{https://BioRender.com/w29a405}).}}
    \label{fig:framework}
\end{figure*}

In this study, we proposed a FM-based large mixture-of-modality-experts model (MOME) which inherits the long-range modeling capability of a transformer-based FM for multiparametric information fusion and could conduct flexibly inference with the design of mixture of experts (Fig. 1, supplementary Fig. 1).
The model was developed and extensively evaluated based on \editorial{a large-scale} multiparametric breast MRI dataset collected from three hospitals in China, achieving comparable malignancy differentiating performance with National Health Commission (NHC)-qualified radiologists.
With generalizability across data collected from the north, southeast, and southwest of China, the model exhibits its clinical value in decreasing unnecessary biopsies for BI-RADS 4 patients.
Moreover, MOME could also conduct subtyping of triple-negative breast cancer and predicting of response to neoadjuvant chemotherapy.
With these capabilities, we exemplified the clinical value of MOME in non-invasive, personalized management of breast cancer patient.

\begin{figure}[!t]
\centering
  \includegraphics[width=0.9\linewidth]{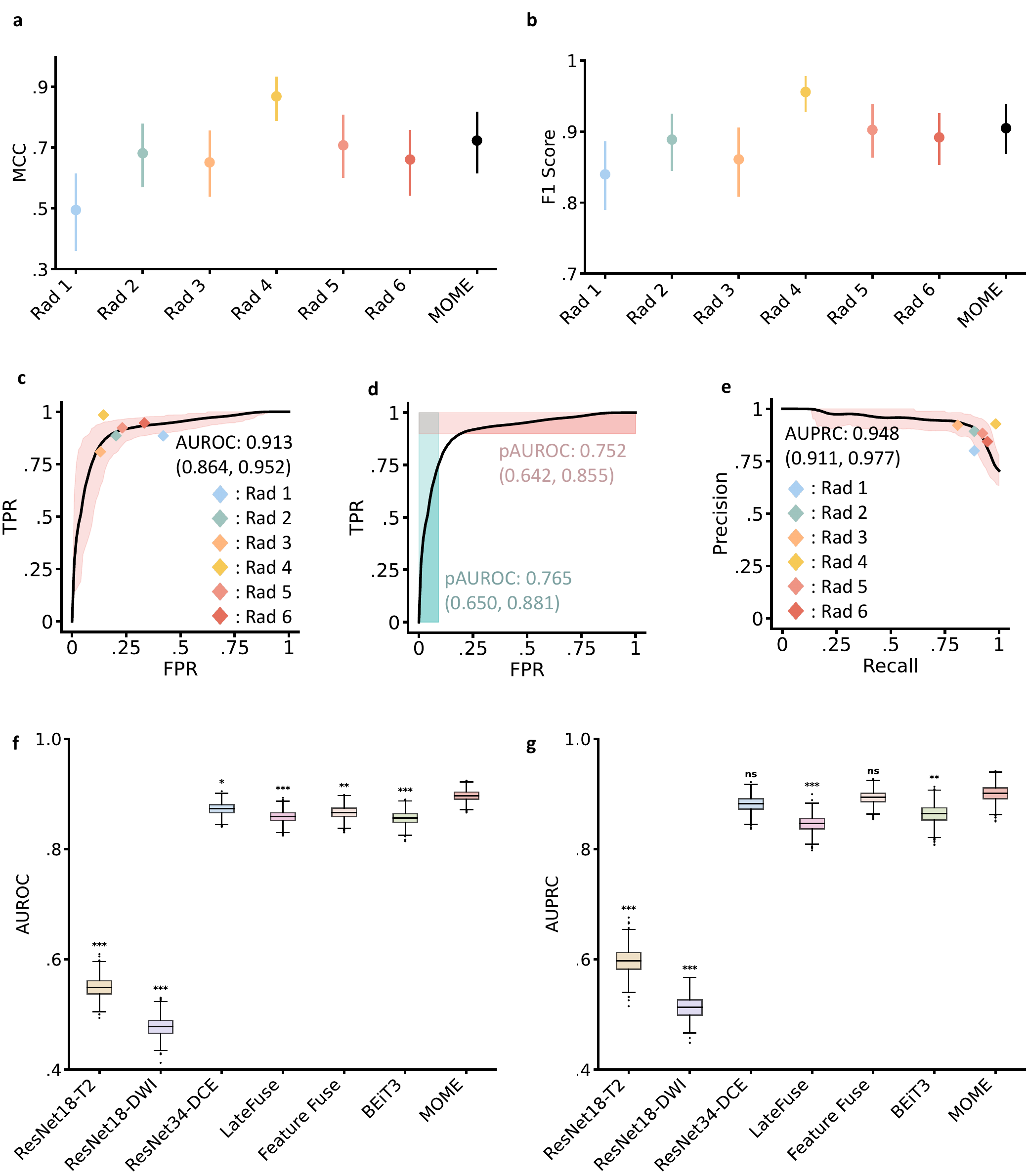}
\caption{\textbf{Discriminative malignancy detection performance of MOME.} MOME achieved comparable MCC (a) and F1 (b) score performance to four experienced radiologists out of six readers, and significantly outperformed one junior radiologist, on the internal testing set 2 (n=200). MOME also showed high AUROC (c), pAUROC (d), and AURPC (e), on the internal testing set 2 (n=200). {Moreover, MOME outperformed other unimodal or multimodal methods in both AUROC (f) and AUPRC (g), on the the combination of DS1 internal testing set 1 and DS2 (n=1042).} In a-g, performance of all models were presented with CIs based on 1000-time bootstrap. In a, b, c, d, and e, error bars represent the 95\% CIs.
In f and g, the box shows the interquartile range (IQR) containing 50\% of the data, with the bottom edge at Q1 (25th percentile), middle line at the median, and top edge at Q3 (75th percentile); the whiskers extend to the most extreme points within 1.5×IQR beyond the box edges, with points beyond the whiskers representing outliers. P-values were computed by comparing the performance of each method against that of MOME \editorial{using bootstrapping}. Rad = Radiologist; MOME = mixture of modality experts; MCC = Matthews's correlation coefficient; AUROC = area under the Receiver Operating Characteristic curve; AUPRC = area under the precision-recall curve; TPR = True Positive Rate; FPR = False Positive Rate; ***: p-value $\leq$ 0.001; **: p-value $\leq$ 0.01; *: p-value $<$ 0.05; ns: not significant. Source data are provided as a Source Data file.}
\label{fig:curve_complete_modal}
\end{figure}

\section*{Results}\label{sec:results}

\subsubsection*{Dataset characteristics}
This study involved a total of 5,220 multiparametric breast MRI examinations from 5,205 patients, collected from three institutes across ten years (\notsure{supplementary Fig. 2}), that is, Dataset 1 (DS1, \notsure{The Fifth Medical Center of Chinese PLA General Hospital}), Dataset 2 (DS2, Shenzhen People's Hospital), and Dataset 3 (DS3, Yunnan Cancer Hospital).
DS1 comprised \notsure{1,824} examinations took between November 2012 and July 2017.
DS2 comprised \notsure{735} examinations collected between December 2018 and March 2022, and DS3 comprised \notsure{2,661} examinations obtained between November 2015 and October 2022.
For malignancy classification, DS1 was randomly split into training set (n=1,167) and validation set (n=150) for model development, internal testing set 1 (n=307) for evaluation, and internal testing set 2 (n=200) for comparison with radiologists.
Different sets did not have overlapping patients.
DS2 and DS3 were used for external testing. 
In addition, 1,005 subjects and 358 subjects from DS 1 were used for triple-negative breast cancer (TNBC) subtyping and neoadjuvant chemotherapy (NACT) response prediction, respectively, where performance was reported using five-fold cross validation.
More details of patient characteristics can be found in \notsure{supplementary Table 1.}

\subsubsection*{Comparable to radiologists}\label{}
We evaluated the malignancy diagnosis performance of MOME on the internal testing set 2 (n=200) and compared its performance with that of six radiologists (reader 1: less than five years of experience in breast MRI; readers 2 and 3: five to ten years of experience in breast MRI; readers 4, 5, and 6: more than ten years of experience in breast MRI; detailed performance can be found in supplementary Table 3).

In comparison to every single reader (Figs. 2a and 2b, supplementary Table 3), no evidence of statistically significant differences was found between the performance of MOME and those of four radiologists (readers 2, 3, 5, and 6), in terms of both F1 and MCC.
In addition, MOME achieved statistically significantly higher F1 (model-radiologist performance=0.065, 95\% CI 0.019, 0.117) and MCC (model-radiologist performance=0.228, 95\% CI 0.090, 0.384) than reader 1 (F1=0.840 [95\% CI 0.789, 0.886]; MCC=0.495 [95\% CI 0.361, 0.615]).
The performance of MOME was also found to be statistically lower (model-radiologist F1=-0.051 [95\% CI -0.087, -0.019]; model-radiologist MCC=-0.145 [95\% CI -0.240, -0.048]) than that of reader 4 (F1=0.956 [95\% CI 0.927, 0.978]; MCC=0.868 [95\% CI 0.787, 0.933]).
MOME achieved an AUROC of 0.913 (95\% CI: 0.864, 0.952) and an AUPRC of 0.948 (95\% CI: 0.911, 0.977), with five out of six dots representing the performance of radiologists' lying on or under the curves, which also indicates that MOME have similar performance to these radiologists with proper decision thresholds (Figs. 2c and 2e).

\subsubsection*{Outperforming unimodal or other multimodal methods}\label{}
\revise{We combined DS1 internal testing set and DS2 to provide a more comprehensive comparison across different methods with a sample size of 1042. MOME achieved an overall AUROC of 0.896 (95\% CI: 0.876, 0.913) and an AUPRC of 0.901 (95\% CI: 0.873, 0.927), which significantly outperformed all seven compared models in terms of AUROC (Fig. 2f) and was significantly better than five out of seven models regarding AUPRC (Fig. 2g).
Among unimodal methods, the DCE-based model achieved the highest performance with an AUROC of 0.873 (95\% CI: 0.852, 0.895) and an AUPRC = 0.882 (95\% CI: 0.854, 0.909), while severe performance degradation was observed for both the T2WI-based model and the DWI-based model.
Compared to the DCE-based model, multimodal methods other than MOME often showed decreased AUROC and AUPRC, except that Feature Fusion showed an increased AUPRC of 0.893 (95\% CI: 0.868, 0.916).
This improved performance was a result of the consistent improvement achieved by MOME on both DS1 and DS2. 
Specifically, MOME consistently outperformed all unimodal and multimodal approaches on the two datasets, with 0.903 AUROC and 0.941 AUPRC on DS1 internal testing set 1, and external 0.893 AUROC and 0.882 AUPRC on DS2 (Supplementary Table 4).
More detailed values of different metrics on the combined dataset, as well as on DS1 and DS2, can be found in Supplementary Table {\notsure{\revise{4}}}}

These findings indicate that integrating multiparametric MRI information is challenging, while MOME has better capability of multimodal data fusion and classification.

\begin{figure}[!t]
\centering
  \includegraphics[width=0.8\linewidth]{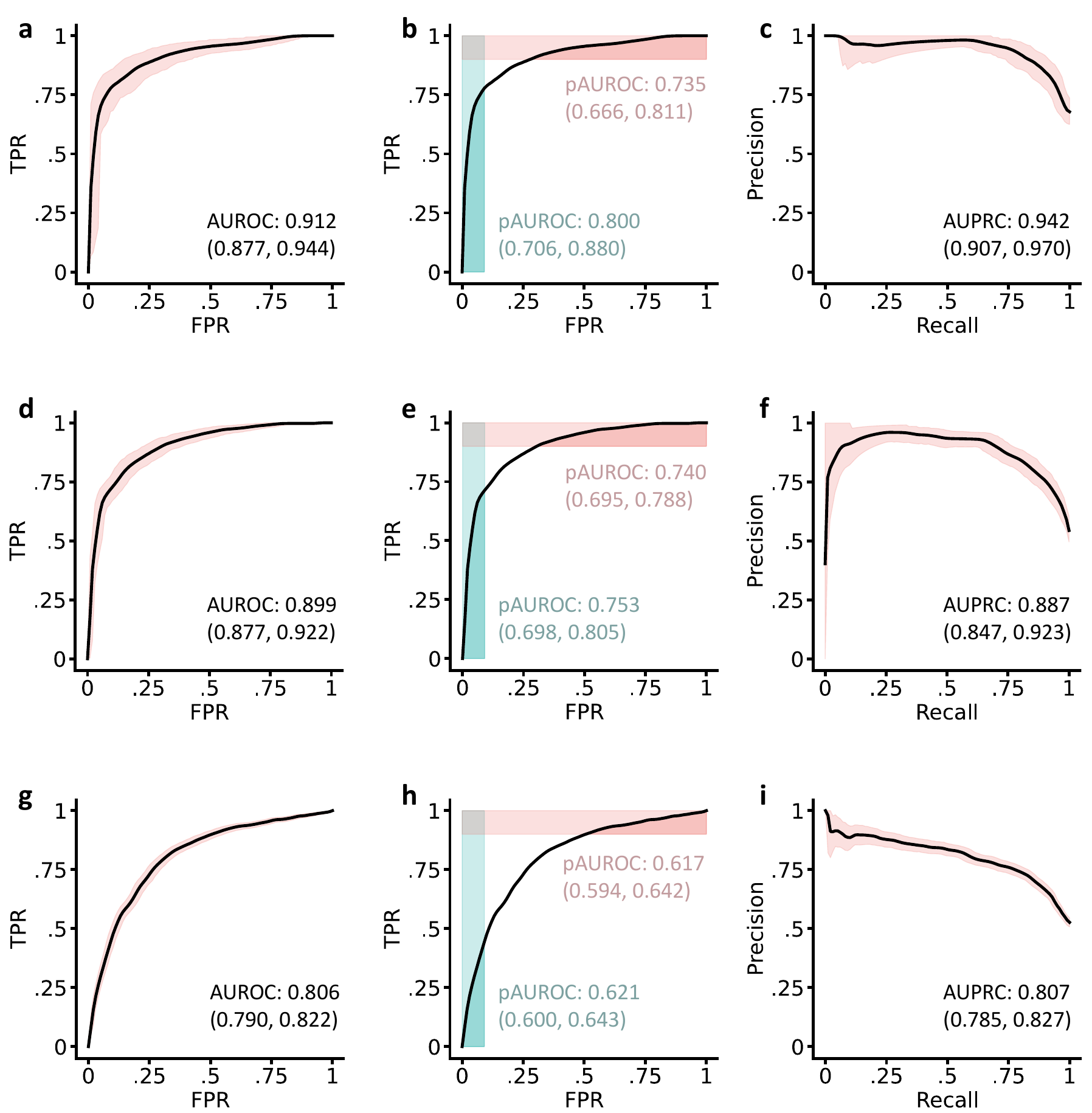}
\caption{\textbf{Malignancy diagnosis performance of MOME across different hospitals.} The results correspond to the ROC curve (a), ROC curve with partial AUC (b), and precision-recall curve (c) on DS1 internal testing set; the ROC curve (d), ROC curve with partial AUC (e), and precision-recall curve (f) on DS2; and the ROC curve (g), ROC curve with partial AUC (h), and precision-recall curve (i) on DS3. The ROCs and precision-recall curves are drawn based on 1000-time bootstrap with 95\% CI. Error bars represents the 95\% CIs. AUROC = area under the Receiver Operating Characteristic curve; AUPRC = area under the precision-recall curve; pAUROC = partial area under the Receiver Operating Characteristic curve; TPR = True Positive Rate; FPR = False Positive Rate. Source data are provided as a Source Data file.}
\label{fig:curve_complete_modal}
\end{figure}

\subsubsection*{Generalizability across hospitals}\label{}
MOME showed generalizable performance on differentiating malignancies from benign tumors across hospitals (Fig. 3).
On the internal testing set 1 (n=307), the ROC (Fig. 3a) and PRC (Fig. 3c) analyses showed that MOME achieved 0.912 AUROC (95\% CI 0.877, 0.944) and 0.942 AUPRC (95\% CI 0.907, 0.970) on the internal testing set (n=307). 
Partial AUROC (pAUROC) at 90\% sensitivity was 0.735 (95\% CI 0.666. 0.811) and 0.800 (0.706 to 0.880) at 90\% specificity (Fig. 3b).
These results indicate that MOME identifies breast cancer patients with a high degree of accuracy.

External validation was conducted on DS2 and DS3.
Except for different breast MRI protocols (Methods section), DS2 and DS3 also possessed different demographics and different distributions of malignant cases compared to the internal dataset.
Specifically, on DS2, MOME achieved 0.899 AUROC (95\% CI 0.877, 0.922) and 0.887 AUPRC (95\% CI 0.847, 0.923).
pAUROC at 90\% sensitivity was 0.740 (95\% CI 0.695. 0.788) and 0.753 (95\% CI: 0.698, 0.805) at 90\% specificity.
On DS3, MOME achieved 0.806 AUROC  (95\% CI: 0.790, 0.822) and 0.807 AUPRC (95\% CI: 0.785, 0.827).
pAUROC at 90\% sensitivity was 0.617 (95\% CI 0.594, 0.642) and 0.621 (95\% CI: 0.600, 0.643) at 90\% specificity.
More detailed performance can be found in \notsure{supplementary Table 5.}
These results reveal that MOME is discriminative and robustness.

\begin{figure}[!t]
\centering
  \includegraphics[width=0.85\linewidth]{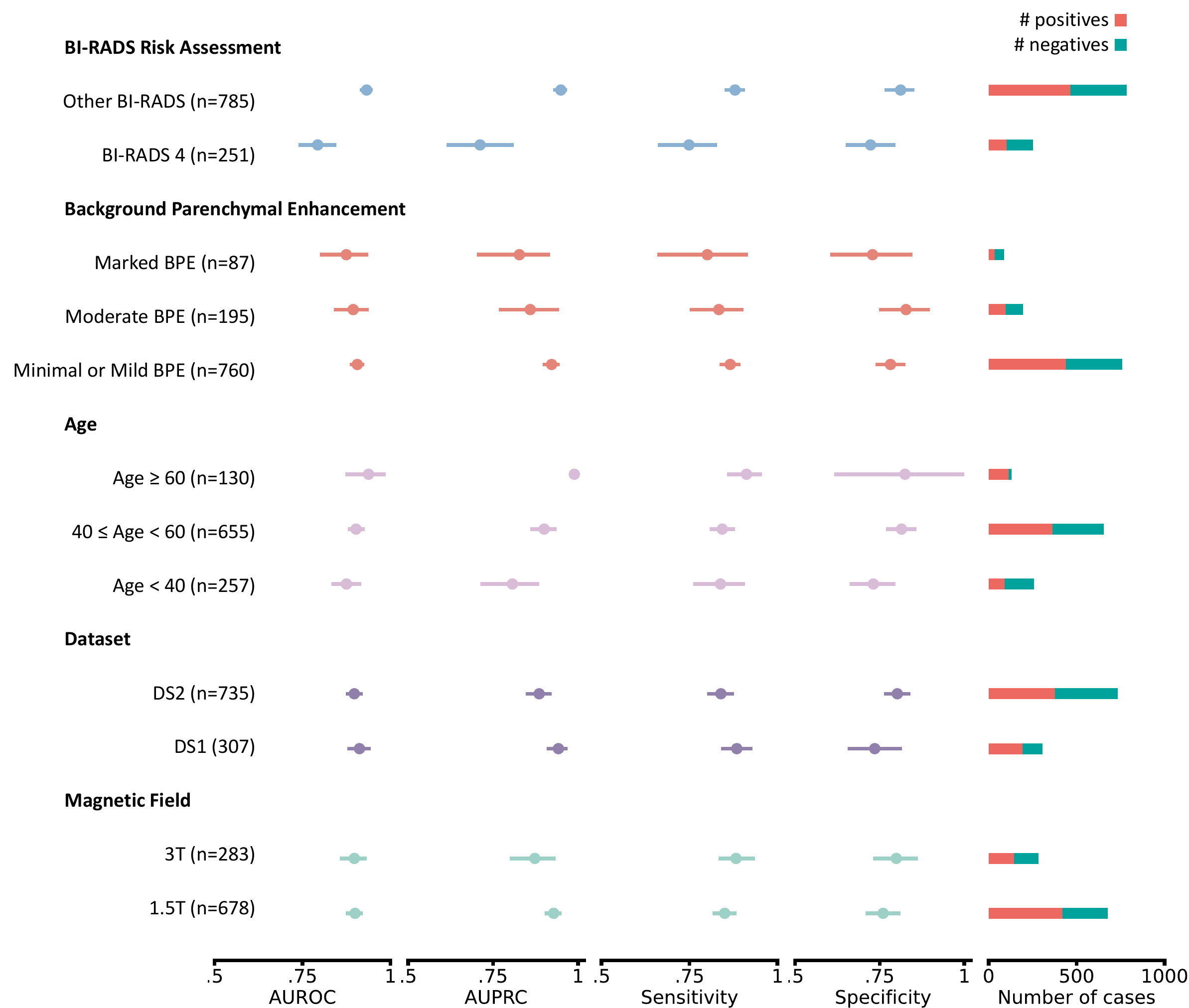}
\caption{\textbf{Malignancy diagnosis performance of MOME on key subgroups on the combination of DS1 test set 1 and DS2.} AUROC, AUPRC, sensitivity, and specificity are reported for each subgroup, from left to right. Red and green bars at the right represent the number of malignant and benign cases for each subgroup. All metrics are presented with 95\% CI based on 1000-time bootstrap. AUROC = area under the Receiver Operating Characteristic curve; AUPRC = area under the precision-recall curve; BI-RADS = Breast Imaging-Reporting and Data System; BPE = Background Parenchymal Enhancement. DS1 = dataset 1; DS2 = dataset 2. Source data are provided as a Source Data file.}
\label{fig:subgroup}
\end{figure}

\subsubsection*{Ablation investigation on modules and missing sequences}\label{}

To investigate the influence of different modules, we developed different variants of MOME by removing each single component (ablation section in Methods) and compared their performance on the DS1 internal testing set 1.
As can be observed from Table 1, removing all of the modality experts led to performance drop in all of the metrics (rows 1).
On the metrics of AUROC and AUPRC, this variants had decreases of 1.0\% and 1.1\%, respectively, compared to those of MOME.
We then replaced the soft mixture of modality experts with a MLP adapter (row 2), leading to performance drops of 2.3\% for AUROC and 3.0\% for AUPRC compared to those of MOME.
All other metrics except for sensitivity also drop compared to the full method (Supplementary Table 6).
These findings indicate that each module contributes to the improved final results.

We also investigated the ability of MOME in inferring with missing sequences.
As DCE remains the most important sequence for breast MRI, we reported the model performance when inferring purely based on DCE (Table 1, supplementary Table 7).
As can be observed, MOME achieved an AUROC of 0.877 and an AUPRC of 0.926.
When adding test-time augmentation, MOME achieved 0.886 AUROC (95\% CI 0.845, 0.920), 0.897 AUROC (95\% CI 0.839, 0.944), 0.881 AUROC (95\% CI 0.858, 0.906), 0.790 AUROC (95\% CI 0.772, 0.806) on internal testing set 1, internal testing set 2, DS2, and DS3 (supplementary Table 7).
These results show that MOME relies on complete multiparametric input for improved diagnosis performance, yet it can also achieve robust inference under the conditions of missing sequence(s).

\begin{figure}[!t]
\centering
  \includegraphics[width=0.8\linewidth]{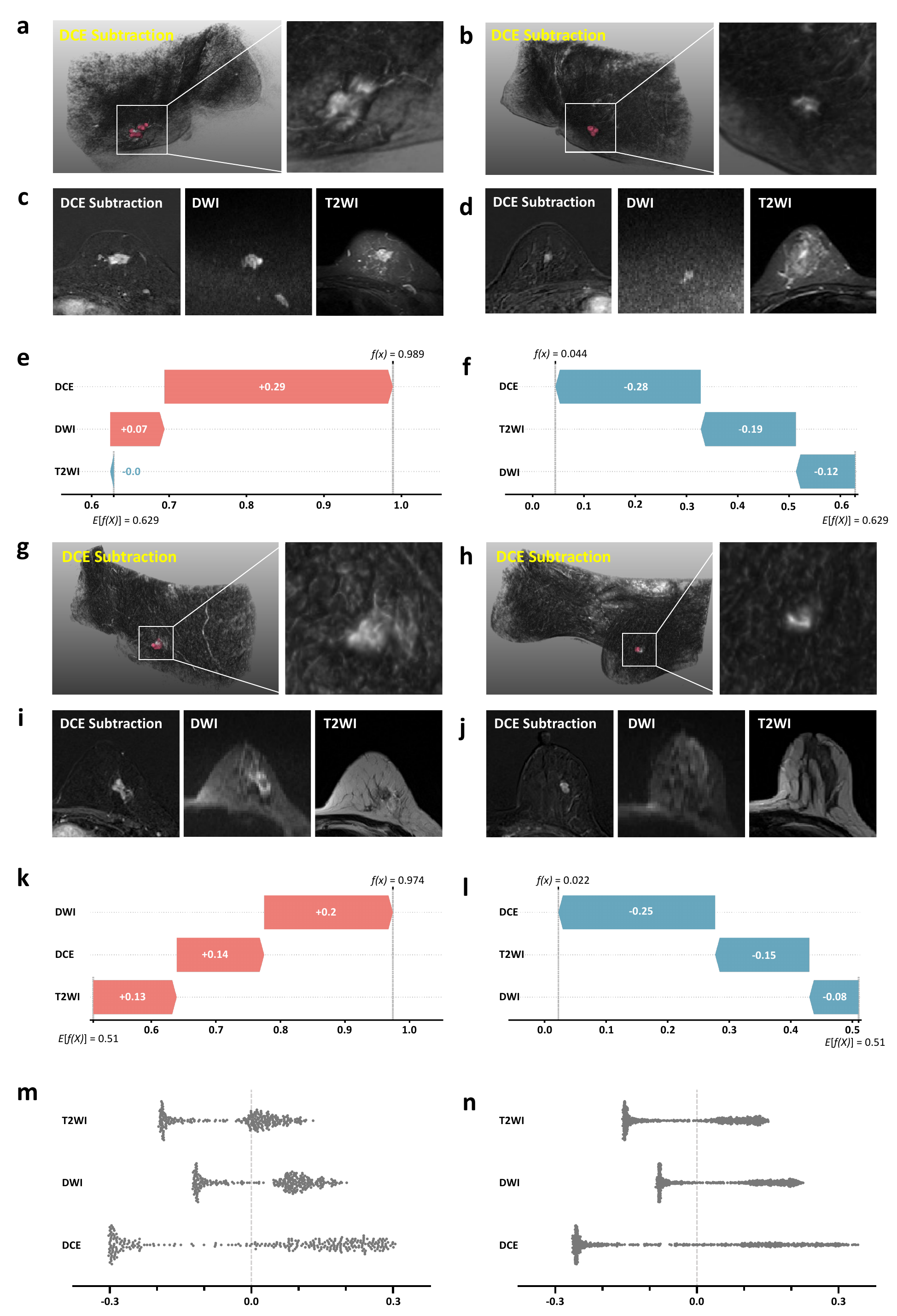}
\caption{\textbf{Decision Interpretation of MOME.} The illustrations correspond to DCE subtraction 3D visualization \revise{overlaid with saliencies in red computed by integrated gradient} (a,b,g,h), the zoomed-in axial view of DCE subtraction, DWI, and T2WI (c,d,i,j), the local Shapley value (e,f,k,l), and global Shapley value of the DS1 internal testing set (m) and DS2 (n). Four typical cases of a BI-RADS 5 patient with a malignant lesion (a,c,e), a BI-RADS 4 patient with a benign lesion (b,d,f) from DS1 internal testing set, and a BI-RADS 5 patient with a malignant lesion (g,i,k), a BI-RADS 4 patient with a benign lesion (h,j,i) from DS2 are shown. BI-RADS = Breast Imaging-Reporting and Data System; DCE = dynamic contrast-enhanced; DWI = Diffusion-weighted Imaging; T2WI = T2-weighted imaging. Source data are provided as a Source Data file.}
\label{fig:interpretation}
\end{figure}

\subsubsection*{Subgroup analysis}\label{}
We conducted subgroup analyses on the combination of DS1 internal test set 1 and DS2, based on several breast cancer risk-related criteria, including the status of background parenchymal enhancement (BPE; the breast tissue enhances on contrast MRI), age, and BI-RADS scores, as well as the magnetic field strength (Fig. 4 and \notsure{supplementary Table 8}).
Overall, MOME consistently achieved high AUROC in all groups.

Generally, the model performance was higher on elderly patients (age $\geq$ 60: AUROC = 0.938 [95\% CI 0.872, 0.986], AUPRC = 0.989 [95\% CI 0.975, 0.998]; age between 40 and 60: AUROC = 0.902 [95\% CI 0.878, 0.926], AUPRC = 0.900 [95\% CI 0.859, 0.937]; age $<$ 40: AUROC = 0.875 [95\% CI 0.832, 0.917; AUPRC = 0.807 [95\% CI 0.714, 0.885]).
Trends of increased performance were also observed for women with less BPE (Minimal or Mild BPE: AUROC = 0.906 [95\% CI 0.885, 0.925], AUPRC = 0.922 [95\% CI 0.896, 0.946]; Moderate BPE: AUROC = 0.894 [95\% CI 0.840, 0.939], AUPRC = 0.859 [95\% CI 0.767, 0.945]; Marked BPE: AUROC = 0.875 [95\% CI 0.799, 0.937], AUPRC = 0.828 [95\% CI 0.702, 0.918]).
A noticeable performance drop was found in the BI-RADS 4 group compared to patients with other BI-RADS scores (BI-RADS 4: AUROC = 0.793 [95\% \revise{CI} 0.738, 0.847], AUPRC = 0.712 [95\% \revise{CI} 0.613,0.811], sensitivity = 0.749 [95\% \revise{CI} 0.660,0.828], specificity = 0.723 [95\% \revise{CI} 0.649,0.797]; Other BI-RADS: AUROC = 0.932 [95\% \revise{CI} 0.913,0.949], AUPRC = 0.950 [95\% \revise{CI} 0.927,0.968], sensitivity = 0.879 [95\% \revise{CI} 0.850,0.908], specificity = 0.812 [95\% \revise{CI} 0.765,0.853]).
The model performance was different on DS1 and DS2 (DS1: AUROC = 0.912 [95\% CI 0.877, 0.944], AUPRC = 0.942 [95\% CI 0.907, 0.970], sensitivity = 0.884 [95\% CI 0.839, 0.929], specificity = 0.735 [95\% CI 0.655, 0.815] DS2: AUROC = 0.899 [95\% CI 0.877, 0.922], AUPRC = 0.887 [95\% CI 0.847, 0.923], sensitivity = 0.839 [95\% CI 0.801, 0.877], specificity = 0.802 [95\% CI 0.762, 0.842]).
In addition, MOME performed better on 1.5T MRI in terms of AUROC (1.5T: 0.899 [95\% CI 0.874, 0.921]; 3T: 0.897 [95\% CI 0.856, 0.933)]) and AUPRC (1.5T: 0.928 [95\% CI 0.901, 0.952]; 3T: 0.873 [95\% CI 0.800, 0.935]).

\subsubsection*{Model decision interpretation}\label{}
MOME is interpretable in highlighting the lesions and analyzing the contribution of each modality.
\revise{Local interpretation for each case can be analyzed by integrated gradient \cite{sundararajan2017axiomatic} and Shapley Value \cite{shapley1953value} computed for the single case.}
By integrated gradient, it can be observed that that MOME correctly attended to breast lesions when diagnosing malignant (Figs. 5a and 5g) or benign (Figs. 5b and 5h) patients, consistent across DS1 and DS2.
The Shapley value revealed the contributions of each modality to the final prediction (Figs. 5e, f, k, and l).
It can be observed that DCE and DWI played more important roles in recognizing malignant patients, whereas DCE and T2WI showed greater contributions in differentiating benign patients.
\revise{The global interpretation for a more comprehensive analysis of all cases can be derived from the global Shapley value ( Figs. 5m and n). As can be observed,} DCE obtained the highest global contribution for determining malignancies, while DWI and T2WI mostly contributed to diagnosing malignant patients and benign patients, respectively, and this decision-making rule is consistent across DS1 and DS2.

\begin{figure}[!t]
\centering
      \includegraphics[width=\linewidth]{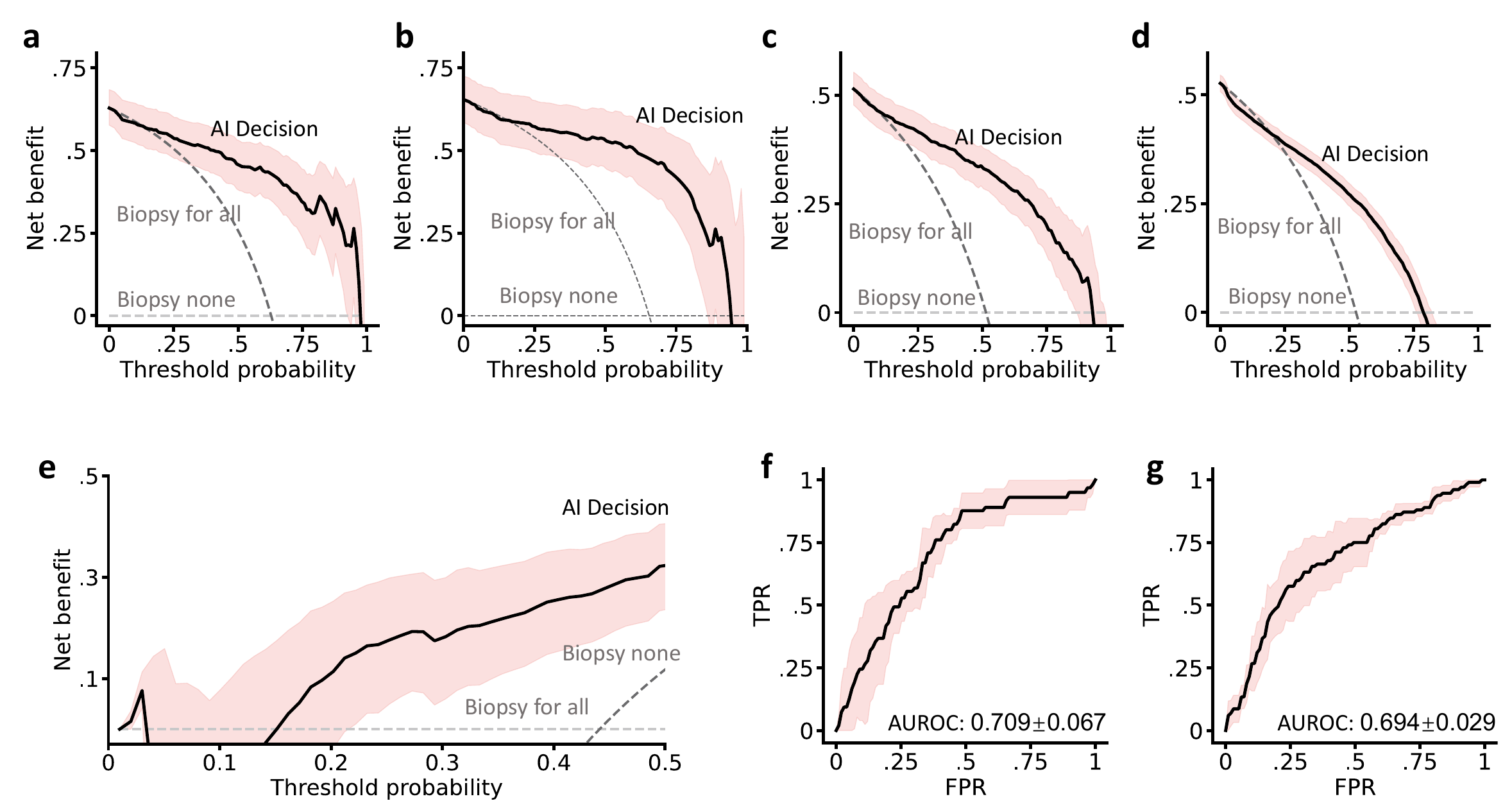}

\caption{\textbf{Potential of noninvasive personalized treatment based on MOME.} The decision curves on DS1 internal testing set 1 (a), DS1 internal testing set 2 (b), DS2 (c), and DS3 (d) show a long range of preference in using MOME for malignancy screening. The decision curve also shows high net benefit of reducing biopsy for BI-RADS 4 patients on DS2 (e). MOME also demonstrated potential with ROC curves for TNBC patient subtyping (f) and NACT response prediction (g). Results are shown with 95\% CIs based on 1000-time bootstrap. Error bars represents the 95\% CIs. TPR = True Positive Rate; FPR = False Positive Rate; Rad = Radiologist. Source data are provided as a Source Data file.}
\label{fig:personalization}
\end{figure}

\subsubsection*{Personalized management}
MOME can be used in clinical process to improve personalized management for breast cancer patients.
We first analyze the net benefit of using MOME to detect patient with malignancies. The decision curve analysis on DS1 internal testing set 1, DS1 internal testing set 2, DS2, and DS3 (Figs. 6a, b, c, d, respectively) indicate high net benefit across long ranges of preference thresholds, demonstrating its potential for decision support in screening malignancies.

MOME can also reduce unnecessary biopsy for BI-RADS 4 patients.
We investigated the trade-off between the number of correctly downgraded cases and the true positive rate by varying operating points of MOME.
Based on the operating point found on DS1 and corresponding results on DS2, we found that at an operating point that identified 7.3\% of BI-RADS 4 patients with benign tumors (8 out of 109 patients), no cancer would be missed with biopsy (n=86).
By decision curve analysis \revise{(Fig. 6e)}, MOME also achieved higher net benefit across a long range of threshold probability, compared to the biopsy all strategy that commonly leveraged for BI-RADS 4 patients. 
\revise{We also compared the performance of reducing unnecessary biopsy among ResNet34-DCE (the best performing single-modality model as shown in Fig. 2), Feature Fuse model (the best performing multimodal model excerpt for MOME as shown in Fig. 2), and MOME, using test-time augmentation. We found that neither ResNet34-DCE or Feature Fuse model could reduce biopsy on DS2 (0 out of 109 patients).}
These results demonstrated a high preference for using MOME for personalized biopsy recommendation for BI-RADS 4 patients. 

In addition, we investigated MOME for triple-negative breast cancer (TNBC) subtyping (Fig. 6f) and NACT pathologic complete response (pCR) prediction (Fig. 6g).
Based on five-fold cross-validation, MOME achieved an AUROC of 0.709$\pm$0.067 for TNBC subtyping (n=1,005) and an AUROC of 0.694$\pm$0.029 for NACT response prediction (n=358).
To note, TNBC patients are also more likely to achieve pCR after NACT.

These results showed that MOME has the potential to facilitate non-invasive personalized breast cancer patient management for malignancy screening, biopsy recommendation, and treatment decision support.

\section*{Discussion}\label{sec12}
The purpose of this work was to show MOME with high differential ability for multiparametric breast MRI analytics with a multiparametric, large-scale, and multi-center study.
Typically, the management of breast cancer patients requires a series of examinations, such as mammography, ultrasound, MRI, histopathology, serology, genomics, and more. 
Personalized management would bring large benefit, yet it depends on accurate malignancy detection, molecular subtyping, and the patients' estimated response to different therapies.
We have shown that MOME could accurately and robustly distinguish patients with malignancies from benign or normal subjects with evaluation across multiple hospitals.
Compared to NHC-qualified radiologists, MOME showed no evidence of statistical differences with the performance of four out of six human experts and statistically significantly higher performance than the junior radiologists with less than five years of breast MRI experience.
Typically, a biopsy is often suggested for a patient with BI-RADS 4 or above after a breast MRI examination, whereas MOME can be used to further characterize BI-RADS 4 patients to alleviate the need of biopsy for the former subgroup.
Moreover, MOME was shown capable of finding TNBC patients who usually have better responses to NACT with an AUROC of 0.709.
Using the pre-treatment multiparametric MRI, we also found that the model could achieve 0.694 AUROC for NACT response prediction.
These capabilities demonstrated MOME's clinical value in biopsy recommendation and treatment decision, thus facilitating efficient, non-invasive, and personalized breast cancer patient management.

Compared to the previous studies, this work presents a more extensive evaluation of a multiparametric deep learning model with large-scale, diverse external assessment for malignancy classification.
Specifically for breast MRI malignancy classification, an AUROC of 0.859 was reported in a single-center study on DCE-MRI from containing 1537 female cases \cite{zhou2019weakly}.
Later, in a multi-center single-sequence study \cite{witowski2022improving} involving 2,2984 DCE-MRI cases, AUROC of 0.92 was reported on its internal testing set.
This model was found to achieve near-perfect performance (AUROCs were above 0.965) on two external datasets (n=922 and 131, respectively) with only invasive breast cancer while exhibiting a lower AUROC of 0.797 on a smaller yet more challenging external set (n=394).
These results were limited to single-sequence MRI, that is, DCE-MRI, and AI-based integration of multiparametric breast MRI is less investigated.
Meanwhile, extensive external evaluation was also recommended to comprehensively assess the model's generalizability \cite{bluemke2020assessing,taylor2022uk}.

This study collected \editorial{a large-sacle} multiparametric breast MRI dataset, representing typical populations from the north, southwest, and southeast of China, across a ten-year period.
DS1, DS2 and DS3 were obtained with diverse imaging protocols, such as different scanning matrices, DWI b values, numbers of DCE-MRI sequences, and different demographics.
Although the metrics on DS3 did not match those on DS1 and DS2, we note that DS3 possessed larger data shifts in terms of imaging protocol and \notsure{cohort size}.
MOME achieved high performance to the two external datasets and performed consistently across different subgroups, demonstrating its generalizability.
These findings demonstrate the ability of MOME to integrate high-dimensional multiparametric information in clinical settings.

Methodologically, MOME \editorial{adapts} the powerful foundation models from the natural image domain to a more complex dimension, that is, the three-dimensional, multimodal breast MRI with temporal information in DCE.
In specific, different breast MRI sequences provide different diagnostic information and result in varied breast cancer differentiation abilities, \revise{as observed in Supplementary Table 4}, which explains the difficulty of multiparametric fusion and inference, as other multimodal approaches may not necessarily outperform the pure DCE-MRI-based model.
MOME had most of the parameters shared for each modality, which mimicked a siamese network structure \cite{bromley1993signature} that can effectively model the similarity and differences among the inputs.
Moreover, the first nine layers utilized sparse modality experts and inter-modality self-attention to adequately extract the modality-specific features.
Then, the last few layers adopted the soft mixture of experts and applied self-attention for the holistic multimodal tokens to encourage cross-modal interaction.
The combination of these characteristics finally led to the improved diagnostic accuracy of MOME.
In addition, MOME offered flexible and explainable inference.
The structural design enables MOME to infer with missing sequence(s), which is highly scalable for incomplete imaging protocols.
MOME also exhibits inherent interpretability by providing pixel-level contributions on the input images and showing highlights of the lesions of interest for malignancy detection.
Furthermore, our model provides a way for modality contribution investigation based on Shapley value with its support for missing-modal inference, \revise{which allows us to understand the decision-making in terms of modality contributions both locally for each case and globally refer for comprehensive analysis of all cases}.

We acknowledge several limitations of the current study.
First, we focused on multiparametric MRI, whereas other data, such as mammograms, breast ultrasounds, health records, and demographics, may also be generated during the clinical process and can provide extra insights into the patient's status.
Our future work will scale the study with more available modalities using the proposed unified model structure.
Second, our comparison with radiologists treated the AI system as a stand-alone reader. It would be of great value to see how the AI interpretation can affect the reader's decision and explore the model's role in real-world clinical settings.
Also, the model yielded patient-level results, and future works should extend to more fine-grained diagnoses to deal with patients with multiple lesions and consider intra-tumor heterogeneity.

To summarize, we provided a large-scale, multi-center, multiparametric study using a foundation model for breast MRI analysis, including malignancy diagnosis, \revise{TNBC molecular subtying,} and NACT response prediction.
The multimodal inference matches the routine breast MRI protocol and the reading standard of breast radiologists.
To tackle the difficulties of fusing different MRI sequences, we \editorial{proposed MOME} that used mixture of experts to adapt a foundation model for 3D multiparametric medical image analysis.
MOME demonstrated accuracy and robustness in diagnosing breast cancers with comparable performance to radiologists and showed promise for subtyping and pre-treatment NACT response prediction, providing a groundwork towards noninvasive and personalized management of breast cancer patients.

\section*{Methods}\label{sec13}

\subsubsection*{Ethics approval}
All datasets were collected under institutional review board approval (KYLX2023-163 by Yunnan Cancer Hospital, KY-2022-4-27-1 by The Fifth Medical Center of Chinese PLA General Hospital, and LL-KY-2021679 by Shenzhen People’s Hospital). All data were de-identified before the development of the model. The IRB waived the requirement for individual informed consent due to the retrospective nature of the study.

\subsubsection*{MRI acquisition}

For DS1, the magnetic resonance imaging (MRI) scans were performed on a 1.5T system (Magnetom Espree Pink; Siemens, Munich, Germany) with an 8-channel breast coil. Patients were positioned prone, with both breasts naturally aligned within the coil. Imaging included conventional scans: an axial T1-weighted 3D non-fat-suppressed sequence (TR/TE: 8.7/4.7 ms, matrix: 896$\times$896, slice thickness: 1 mm), T2-weighted fat suppression (TR/TE: 2900/60 ms, matrix: 640$\times$640, slice thickness: 4 mm), and diffusion-weighted imaging (DWI) with b-values of 400, 800, and 1000 s/mm$^2$ (TR/TE: 6200/104 ms, matrix: 236x120, slice thickness: 4 mm). Dynamic contrast-enhanced MRI was conducted using a 3D fat-suppressed VIBE sequence before and 6 times after bolus injection (0.1 mmol/kg gadopentetate dimeglumine, Magnevist, Bayer, Berlin, Germany) at 2 mL/s, followed by a 20-mL saline flush. The examination spanned 7 minutes, with imaging parameters of TR/TE: 4.53/1.66 ms, matrix: 384$\times$384, and slice thickness: 1.0 mm. Images of each phase were subtracted automatically.

For DS2, breast MRI examinations were performed using either a 1.5T Magnetom Avanto or 3.0T Magnetom Skyra magnetic resonance scanner (Siemens Healthineers, Erlangen, Germany) equipped with a dedicated breast coil. Patients were examined in the prone position. The scanning included the following sequences: axial T1-weighted non-fat suppressed images were acquired with TR/TE parameters of 559/12 ms (1.5T) and 6/2.5 ms (3.0T), a matrix of 448 × 448, and slice thickness of 4 mm (1.5T) and 1.6 mm (3.0T); axial T2-weighted images were obtained with TR/TE of 4500/102 ms (1.5T) and 4740/107 ms (3.0T), a matrix of 512 × 512 (1.5T) and 448 × 448 (3.0T), and slice thickness of 4 mm.
A single-shot echo planar imaging pulse sequence was used to acquire diffusion-weighted images with the following parameters: TR/TE of 6400/97 ms (1.5T) and 5700/59 ms (3.0T), a matrix of 192 × 192 (1.5T) and 340 × 170 (3.0T), slice thickness of 4 mm, and b-values of 50/500/1000 s/mm2 (1.5T) and 50/400/800 s/mm$^2$ (3.0T); DCE-MRI was performed during intravenous injection of 15 ml Gd-DTPA at 0.1 mmol/kg over 6 minutes and 41 seconds at a rate of 2.5 ml/s. The sequence included one pre-contrast axial image and five post-contrast axial scans spaced 30 seconds apart. DCE parameters were: TR/TE of 5.2/2.4 ms (1.5T) and 4.7/1.7 ms (3.0T), a matrix of 384 × 384 (1.5T) and 448 × 448 (3.0T), and slice thickness of 1.1 mm (1.5T) and 1.6 mm (3.0T). Images from each phase were automatically subtracted.

For DS3, breast MRI was performed on a 1.5T system (Magnetom Avanto; Siemens, Germany), equipped with an 4-channel breast phased array surface coil. Patients were examined in the prone position. The scanning steps are as follows: axial T1WI fast low angle shot 3D, flash 3D sequence (TR 8.6 ms, TE 4.7 ms, slice thickness 1 mm); Fat-suppressed transverse axial T2WI rapid inversion recovery (Turbo Inversion Recovery Magnitude, TIRM) sequence (TR 5600 ms, TE 56 ms, slice thickness 4 mm) scan; Axial diffusion-weighted imaging uses single shot echo plannar imaging (SS-EPI) sequence (TR 4900 ms, TE 84 ms, FOV 340 mm, slice thickness 4mm, the diffusion sensitivity factor b value is selected to be 0 s/mm$^2$ and 800 s/mm$^2$). Dynamic contrast-enhanced magnetic resonance imaging (DCE-MRI) scan in transverse position: first scan the first-stage transverse position fat-suppressed T1WI (i.e., masked film), then inject contrast agent, and then continuous scanning of 5 continuous dynamic enhancement sequences, each period is 60 seconds. Twenty ml of Gd-DTPA-BMA (OmniScan, GE Healthcare, Ireland) was injected at a rate of 2.0 ml/s and then flushed with 20 ml of saline. Parameters of DCE-MRI were: TR 4.43 ms, TE 1.5 ms, matrix 352 × 324, slice thickness 1.7 mm. Images of each phase were subtracted automatically. 

A summary of scan parameters for all sequences can be found in \notsure{supplementary Table 2.}

\subsubsection*{Data preprocessing}
We generated three-dimensional breast region masks from T1-weighted fat-suppressed magnetic resonance images, following Zhou et al. \cite{zhou2019weakly}, \revise{to remove irrelevant information and reduce the computational cost for MOME inference}.
Specifically, two-dimensional binary breast masks were obtained for each MR image slice by extracting boundaries and applying morphological processing methods. 
Then, all the two-dimensional masks were stacked to create a three-dimensional mask, which was smoothed using a 3D Gaussian filter ($\gamma=20$).
The obtained 3D masks were used to crop the MRI scans and mask out the air and chest regions.
The main purpose of the breast mask was to reduce the input data dimensions.

%For DS1, we utilized the 3D mask to crop all DCE-MRI subtraction sequences, resizing both DCE-MRI and the corresponding 3D mask to 384×384×128. 
%The breast region in DCE-MRI was then extracted and normalized, followed by element-wise multiplication with the 3D mask. 
\revise{For DS1, we have one pre-DCE T1WI and 6 time points of DCE-MRI, and we took the DCE-MRI subtraction as the input to the model which enhances the tumor regions.
we first resized the 3D mask to be the same shape as the DCE-MRI images (384×384×128 in our dataset). We then utilized the 3D mask to crop all DCE-MRI subtraction sequences (the resulting dimensions are typically smaller than 384×256×128 with a channel dimension of 6), followed by a normalization to linearly scale the voxels to zero mean and unit variance. During training, all cropped MRI scans were again padded to be of the same size of 384×384×128.}
For T2 images, they were first resized to 384×384×32 and then the breast area was cropped, with standardization (linear scaling to zero mean and unit variance) applied subsequently. 
For DWI, the sequence with the highest b value (1000 or 800) was used, and only standardization was performed for pre-processing. 
During training and testing, DCE-MRI, T2WI, and DWI was padded to $384\times256\times128$, $384\times256\times48$,and $256\times128\times32$, respectively.

For DS2 and DS3, the procedures were similar to DS1 except that their DCE-MRI had 5 phases and were interpolated to 6 phases using first-order B-spline interpolation with a grid-constant mode.

\subsubsection*{Groundtruth}
% malignancy
The malignant or benign labels for all patients from DS1, DS2, and DS3 were confirmed by histopathological examination. 
In DS1, 365 patients with breast cancer confirmed by histopathology underwent NACT. One cycle of NACT lasted for 21 days. After the second cycle, clinicians evaluated the response and tolerability of NACT for each patient. All patients underwent MRI scans before treatment and at least 2 follow-up studies. All patients underwent definitive surgery after the final cycle of treatment.
For confirmed breast cancer, molecular subtypes were determined based on the Chinese Anti-Cancer Association and the immuno-histochemical results in the histopathological reports were analyzed by pathologists. The estrogen receptor (ER), progesterone receptor (PR), HER2 status, and Ki-67 index were used to define the molecular subtypes. Estrogen receptor and PR positivity were defined as more than 1\% staining \cite{rakha2017molecular}. HER-2 positivity was defined as a score of 3+ by IHC or fluorescence in situ hybridization amplification with a score of 2+ or higher \cite{rakha2017molecular}. TNBC was determined by ER negative, PR negative, and HER2 negative.

\subsubsection*{MOME}
MOME presents a unified, easily-extendable structure for multimodal data integration, such as multiparametric breast MRI.
The input data were first embedded into features using different tokenizers and concatenated as input to a transformer architecture adapted from a foundation model. 

Each MRI sequence leveraged a tokenizer module with the same structure to embed the input into tokens, which was a sequence of embedded features, denoting as $X^{\rm DCE}$, $X^{\rm DWI}$, and $X^{\rm T2}$.
The tokenizer contained three 3D convolutional layers (stride = 2) and ended with a maxpooling layer (stride = 2). 
Each convolution layer was appended with an instance normalization \cite{ulyanov2016instance} layer and a ReLU layer \cite{nair2010rectified}.
The tokenizer downsampled the width, height, and number of slices of the MRI input with a scale of 1/16 and generated a feature map with 768 feature dimensions.
The obtained feature maps were then flattened at the width, height, and slice dimension to form the input as a sequence.
For example, the DCE-MRI with original shape of $\mathbb{R}^{384 \times 256\times 128 \times 6}$ would be processed \revise{by the tokenizer} into $\mathbb{R}^{24 \times 16\times 8 \times 768}$ and flattened to $\mathbb{\rm X}^{\rm DCE}\in\mathbb{R}^{3072 \times 768}$.
A CLS token $\mathbb{\rm X}^{\rm CLS}\in\mathbb{R}^{1\times768}$ was appended to $\mathbb{\rm X}^{\rm DCE}$ and would be used for final classification.
Finally, a learnable 1d positional embedding was added to the input tokens. \revise{The multimodal feature of different sequences were then concatenated and input to the latter transformer structure.}

The transformer structure of MOME was adapted from BEiT3 \cite{wang2023image}.
BEiT3 was originally a vision-language foundation model with 276 million parameters and was pre-trained from 21 million image-text pairs, 14 million images, and 160 gigabytes of documents.
The model contains twelve transformer blocks with the same structure for feature encoding, of which the process can be formulated as follows:
\begin{align}
    \mathbb{\rm Z}^{l} = \mathbb{\rm X}^{l} + MSA((LN(\mathbb{\rm X}^{l}))), \\
    \mathbb{\rm X}^{l+1} = \mathbb{\rm Z}^{l} + FFN(LN(\mathbb{\rm Z}^{l})),
\end{align}

where $\mathbb{\rm X}^{l}$ is the input to the $l$-th block, MSA stands for the multi-head self-attention, FFN represents a feed-forward network with two linear projection layers, LF means layer norm \cite{ba2016layer}, and $\mathbb{\rm X}^{l+1}$ is the output and will be fed to the $l+1$-th transformer block.

Based on the foundation model, BEiT3, we further introduced the mixture of modality experts  (MOME)to enable multimodal learning and fusion.
Specifically, we fixed the pre-trained parameters of BEiT3 and injected simple trainable modules.
We set the first k layers to learn from each different modality (that is, each different MRI sequence) by adding the sparse mixture of modality experts into the transformer blocks, which can be formulated as follows:
\begin{align}
    &\mathbb{\rm Z}^{l}_{i} = \mathbb{\rm X}^{l}_{i} + MSA^{*}((LN^{*}(\mathbb{\rm X}^{l}_{i}))), \\
    &\mathbb{\rm X}^{l+1}_{i} = \mathbb{\rm Z}^{l}_{i} + FFN^{*}(LN^{*}(\mathbb{\rm Z}^{l}_{i})) + MOME^{\rm Sparse}_{i}(LN^{*}(\mathbb{\rm Z}^{l}_{i})),
\end{align}
where * means that the parameters of the module were loaded from pre-trained BEiT3 and fixed during training, and $i$ is an index.
In particular, $\mathbb{\rm X}^{l}_{i}$ is the $i$-th feature from the set $\{\mathbb{\rm X}^{\rm DCE,l}, \mathbb{\rm X}^{\rm DWI,l}, \mathbb{\rm X}^{\rm T2,l}\}$, and $MOME^{\rm Sparse}_{i}$ is the $i$-th sparse MOME.
Following the structure proposed by Yang et al. \cite{yang2022aim}, $MOME^{\rm Sparse}$ takes the structure of a feed-forward structure with two layers of linear projections followed by a layer norm:
\begin{equation}
    MOME^{\rm Sparse}(\mathcal{X}) = LN(LP(GELU(LP(\mathcal{X})))),
\end{equation}
where LP stands for a linear project layer, and GELU means the Gaussian Error Linear Units \cite{hendrycks2016gaussian}.
In this way, each sparse MOME would learn to encode one specific type of input sequence while most of the transformer block parameters were fixed and shared for each modality.

The original BEiT3 has no knowledge of fusing the multiparametric MRI information.
Therefore, the multimodal fusion was conducted by the last 12-k transformer blocks using soft MOME, that is, $MOME^{Soft}$.
Here, the embedded features were concatenated to be the input to the adapted foundation model, denoted as $X=[\mathbb{\rm X}^{\rm DCE}, \mathbb{\rm X}^{\rm DWI}, \mathbb{\rm X}^{\rm T2}]$, and a transformer block is formulated as follows:

\begin{align}
    &\mathbb{\rm Z}^{l} = \mathbb{\rm X}^{l} + MSA^{*}((LN^{*}(\mathbb{\rm X}^{l}))), \\
    &\mathbb{\rm X}^{l+1} = \mathbb{\rm Z}^{l} + FFN^{*}(LN^{*}(\mathbb{\rm Z}^{l})) + MOME^{\rm Soft}(LN^{*}(\mathbb{\rm Z}^{l})),
\end{align}

and the formulation of $MOME^{Soft}$ can be further elaborated as follows:
\begin{equation}
    MOME^{Soft}(\mathcal{X}) = LN(LP(SMoE(GELU(LP(\mathcal{X}))))),
\end{equation}
where $SMoE$ is a soft mixture of expert (Soft MoE) \cite{puigcerver2023sparse}.
Here, the linear projection layers were also anticipated to reduce and expand the feature dimensions, and $SMoE$ was expected to learn to integrate the multimodal information.

Without loss of generation, let $\mathcal{X}\in\mathbb{R}^{m\times d}$ be the input to $SMoE$, where $m$ is the number of tokens of the input sequence data, and $d$ is the feature dimension.
In detail, soft MoE first obtains a set of slots $\tilde{\mathcal{X}}$ that are linear combinations of all of the input tokens.
The process can be further elaborated as follows:
\begin{align}
    &D_{i,j} = \frac{exp((\mathcal{X}\Phi)_{i,j})}{\sum_{i'=1}^{m}exp((\mathcal{X}\Phi)_{i',j})}, \\
    &\Tilde{\mathcal{X}} = D^{T}\mathcal{X},
\end{align}
where $\Phi\in\mathbb{R}^{d\times(n\cdot p)}$ is a learnable linear projection.
By the above, $n\cdot p$ slots were generated, and every $p$ slots will be processed by an expert function, resulting in totally $n$ experts.
Let $f$ represent the expert function, the rest process of soft MoE is to conduct expert function over the slots and map the slots back to tokens:
\begin{align}
    &\Tilde{\mathcal{Y}_i}=f_{\lfloor i/p\rfloor}(\Tilde{\mathcal{X}_i}),\\
    &C_{i,j} = \frac{exp((\mathcal{X}\Phi)_{i,j})}{\sum_{j'=1}^{m}exp((\mathcal{X}\Phi)_{i,j'})},\\
    &\mathcal{Y} = C\Tilde{\mathcal{Y}},
\end{align}    
where $\Tilde{\mathcal{Y}}\in\mathbb{R}^{(n\cdot p)\times d}$, and the expert function $f$ is a linear projection function.
The multiple expert functions were aimed to learn different fused features from the long sequence input generated from the multiparametric MRI.
Then, the combination of experts acted as an ensemble learning strategy to improve the fusion results.

After twelve transformer blocks, the CLS token was extracted and fed into a linear classification layer after layer norm.
For inference with missing modalities, we simply removed the sparse expert corresponding to the missing sequence.

\subsubsection*{Compared Methods}

Late Fusion took the average of the three unimodal models' outputs
Feature Fusion concatenated the features from the unimodal models and generated output using a three-layer perception.
The comparison with BEiT3 was to show that directly using the pre-trained parameters was not enough to utilize the foundation model's capability. We took the vision transformer part of BEiT3, fixed its parameters, and concatenated the multiparametric features to be the inputs, where the multimodal fusion was then conducted by the pre-trained self-attention modules inside BEiT3.

\subsubsection*{Implementation details}
The first 9 transformer blocks of MOME were implemented with sparse MOME, and the last 3 transformer blocks of MOME were implemented with soft MOME.
The number of experts used in soft MOME was set to 128, and each expert processed one input slot.
Adam optimizer \cite{kingma2014adam} was used with an initial learning rate of $1\times 10^{-4}$ gradually reduced to $1\times 10^{-6}$ using a cosine annealing strategy \cite{loshchilov2016sgdr}.
For malignancy classification, each model was trained with a batch size of 1 in 100 epochs and under the supervision of the standard softmax function.
Results for this task were obtained after test-time augmentation, except when compared with other models.
For TNBC subtyping and NACT response prediction, the model parameters were initialized from those trained on the malignancy classification task and trained with 200 epochs.
The result for NACT response prediction was obtained after test-time augmentation.
The weighted softmax function was used for these two tasks, of which the weights were determined by the ratio of positive and negative samples.
All implementations was based on PyTorch \cite{paszke2019pytorch} with an NVIDIA GeForce RTX 3090 GPU.

During training, data augmentation including random padding and random one-axis or two-axis flipping was adopted to enhance the data diversity.
During inference, MOME generated patient-level prediction in the range of (0, 1). 
The checkpoints with the best AUROCs on the validation set were used.
For test-time augmentation, we generated nine different padded versions of the input 3D MRI with six different flipping situations, resulting in 54 different augmented versions.
For test-time augmentation, the averaged model scores on the 54 augmentations were used as the final results.
A score larger than 0.5 was considered a positive case (for example, a patient with breast cancer), otherwise it was considered a negative case (for example, a benign case).
For evaluation on the trade-off between the correctly downgraded cases and the true positive rate, we first varied the operating point on the DS1 internal testing set1 and reported the results on DS2 using the obtained threshold.

For model interpretation, integrated gradient \cite{sundararajan2017axiomatic} was used to find the highlighted regions on the MRI images. \revise{MeVisLab (\url{https://www.mevislab.de/}) was used to display the 3D visualization with integrated gradient in Fig. 5.}
Shapley values were computed by inferencing with all seven combinations of the three modalities.

\subsubsection*{Reader study}
Six NHC-qualified radiologists with varying levels of experience (one less than 5 years, two 5-10 years, and three over 10 years) participated in an online reader study. The study involved the evaluation of 200 breast MRI examinations using the BI-RADS classification system. Readers were provided with the MRI images, including T2WI, T1WI, DWI, and DCE-MRI, and had no access to ADC values, and time-intensity curves (TIC). The readers were also blinded to the clinical history and pathology results, providing BI-RADS classifications based solely on the image information.
When compared to the models, a BI-RADS of 4b or above was considered that the reader made a diagnosis of a malignancy.
If at least one malignant lesion was detected, the patient-level diagnosis was considered malignant, otherwise negative.

\subsubsection*{Statistical analysis}
To measure model performance, the area under the receiver operating characteristic (AUROC), the area under the precision-recall curve (AUPRC), partial AUROC under 90\% sensitivity and 90\% specificity, accuracy, F1-score, sensitivity, specificity, the positive predictive values (PPV), and the negative predictive values (NPV), and the Matthews's correlation coefficient (MCC) were used to provide comprehensive evaluations.
All results were reported with 95\% CIs with 1000-time bootstrap \cite{carpenter2000bootstrap}.  \revise{P-values were computed based on bootstrapping (\url{https://gist.github.com/rajpurkar/f96c131ba3aeffb1927255d4363496a9}).}

Two scenarios were considered for decision curve analysis \cite{vickers2006decision} to estimate the net benefit of taking a treatment (that is, surgery, as in Figs. \revise{6a, 6b, 6c, and 6d}) or not taking an intervention (that is, biopsy, as in \revise{Fig. 6e}).
The reasons are that, when considering the whole cohort, discovering cancer patients is essential, and when considering only the BI-RADS 4 patients, finding out those who can avoid biopsy would be of more value.
We report the decision curves with 95\% CIs using a 1000-time bootstrap to avoid overestimation.

All measurements are implemented with Python 3.9.

\subsubsection*{Illustrations}
Multiple elements in Figure 1 were created with BioRender.com.
In Figure 1, the woman on breast MRI illustration was from ST.art/Shutterstock.com.
Most plots were created using the matplotlib library in Python. \revise{The 3D visualization of DCE-MRI in Fig. 5 was obtained using MeVisLab (\url{https://www.mevislab.de/}).}

\subsubsection*{Data availability}
\editorial{The raw in-house data from DS1, DS2, and DS3 are protected and are not publicly available due to data privacy, while supporting the findings including the imaging data can be available under restricted access for non-commercial and academic purposes only.
Access can be obtained by request to the corresponding authors.} The requirements will be evaluated concerning institutional policies, and data can only be shared for non-commercial academic usage with a formal material transfer agreement. All requests will be promptly reviewed within a timeframe of 20 working days. Source data are provided in this paper.

\subsubsection*{Code availability}
\editorial{The complete source code for this study is available at \url{https://github.com/LLYXC/MOME/tree/main}. The software is licensed under the Apache License 2.0. Our implementation incorporates code from two external sources: 1. BEiT-3 (\url{https://github.com/microsoft/unilm/tree/master}) by Microsoft Corporation, which is licensed under the MIT License. 2. soft-moe (\url{https://github.com/bwconrad/soft-moe/tree/main}), which is licensed under the Apache License 2.0. No additional restrictions beyond those specified in the Apache License 2.0 apply to our code.}
\backmatter

\bibliography{sn-article}% common bib file

\section*{Acknowledgments}
This work was supported by the Innovation and Technology Commission (Project No. MHP/002/22 and ITCPD/17-9), received by H. Chen, Shenzhen Science and Technology Innovation Committee Fund (Project No. SGDX20210823103201011), received by H. Chen, Research Grants Council of the Hong Kong Special Administrative Region, China (Project No. T45-401/22-N), received by H. Chen, National Natural Science Foundation of China (Project No. 82102142),  received by J. Zhou, First-Class Discipline Team of Kunming Medical University (Project No. 2024XKTDYS08, 2024XKTDTS06 and 2024XKTDTS03), received by Z. Li, and National Key R\&D Program of China (Project No. 2023YFE0204000),  received by H. Chen. We thank Dr. Herui Yao from Sun Yat-sen Memorial Hospital, Sun Yat-sen University, Guangzhou, China for supporting this study.

\section*{Author contributions}
L. Luo, H. Chen, Z. Li, and J. Zhou conceived the study. L. Luo planned and executed the experiments and analytics. M. Li, M. Wu, Z. Li, and J. Zhou collected and labeled the multiparametric data. L. Luo and Y. Xin conducted pre-processing for the data. L. Luo, Z. Li, J. Zhou, P. Rajpurkar, and H. Chen drafted the manuscript. All authors (L. Luo, M. Wu, M. Li, Y. Xin, Q. Wang, V. Vardhanabhuti, W. CW Chu, Z. Li, J. Zhou, P. Rajpurkar, and H. Chen) provided critical feedback for the manuscript. All authors read and approved the manuscript.

\section*{Competing Interests Statement}
\editorial{The Authors declare no competing interests.}

\section*{Tables}

\begin{table}[!h]
\renewcommand{\arraystretch}{1.3}
    \begin{tabular}{cccc}
    \hline
    Model & Inference & AUROC & AUPRC \\
    \hline
    w/o MOME & Multiparametric & 0.893 (0.852, 0.929) & 0.930 (0.891, 0.962) \\
    w/o MOME$^{\rm Soft}$ & Multiparametric & 0.880 (0.837, 0.920) & 0.911 (0.864, 0.951) \\
    \multirow{2}{*}{MOME} & DCE & 0.877 (0.836, 0.913) & 0.926 (0.895, 0.951) \\
    & Multiparametric & 0.903 (0.866, 0.936) & 0.941 (0.910, 0.965) \\
    \hline
    \end{tabular}
    \caption{AUROC and AUPRC performance in ablation study on MOME. Results are reported as the mean and 95\% CI with 1000-time bootstrap. w/o MOME: all modality experts are removed. w/o MOME$^{\rm Soft}$: the soft mixture of experts at the last three layers were removed. Results are obtained without test-time augmentation.}
    \label{tab:ablation_auroc_auprc}
\end{table}

\section*{Figure Captions}
Fig. 1. \textbf{Overview of the multiparametric breast MRI-based study design.} a. Data collection from three different hospitals, covering the population of the north, southeast, and southwest of China. b. The generation of multiparametric breast MRI, where T2-weighted MRI, Diffusion-weighted MRI, and DCE-MRI were mainly used in this study. c. MOME first takes multi-parametric MRI as input. Then, based on pre-trained foundation model parameters, mixture of sparse modality experts and soft modality experts are leveraged for unimodal feature extraction and multimodal information integration. d. MOME can be used for malignancy screening, molecular subtyping, and NACT response prediction, offering non-invasive personalized management for breast cancer patients. DS1 = dataset 1; DS2 = dataset 2; DS3 = dataset 3; MOME = mixture of modality experts; DCE = dynamic contrast-enhanced; DWI = diffusion-weighted Imaging; NACT = neoadjuvant chemotherapy. \editorial{Multiple elements were created with BioRender.com (\url{https://BioRender.com/w29a405}).}

Fig. 2. \textbf{Discriminative malignancy detection performance of MOME.} MOME achieved comparable MCC (a) and F1 (b) score performance to four experienced radiologists out of six readers, and significantly outperformed one junior radiologist, on the internal testing set 2 (n=200). MOME also showed high AUROC (c), pAUROC (d), and AURPC (e), on the internal testing set 2 (n=200). {Moreover, MOME outperformed other unimodal or multimodal methods in both AUROC (f) and AUPRC (g), on the the combination of DS1 internal testing set 1 and DS2 (n=1042).} In a-g, performance of all models were presented with CIs based on 1000-time bootstrap. In a, b, c, d, and e, error bars represent the 95\% CIs.
In f and g, the box shows the interquartile range (IQR) containing 50\% of the data, with the bottom edge at Q1 (25th percentile), middle line at the median, and top edge at Q3 (75th percentile); the whiskers extend to the most extreme points within 1.5×IQR beyond the box edges, with points beyond the whiskers representing outliers. P-values were computed by comparing the performance of each method against that of MOME \editorial{using bootstrapping}. Rad = Radiologist; MOME = mixture of modality experts; MCC = Matthews's correlation coefficient; AUROC = area under the Receiver Operating Characteristic curve; AUPRC = area under the precision-recall curve; TPR = True Positive Rate; FPR = False Positive Rate. Source data are provided as a Source Data file.

Fig. 3. \textbf{Malignancy diagnosis performance of MOME across different hospitals.} The results correspond to the ROC curve (a), ROC curve with partial AUC (b), and precision-recall curve (c) on DS1 internal testing set; the ROC curve (d), ROC curve with partial AUC (e), and precision-recall curve (f) on DS2; and the ROC curve (g), ROC curve with partial AUC (h), and precision-recall curve (i) on DS3. The ROCs and precision-recall curves are drawn based on 1000-time bootstrap with 95\% CI. Error bars represents the 95\% CIs. AUROC = area under the Receiver Operating Characteristic curve; AUPRC = area under the precision-recall curve; pAUROC = partial area under the Receiver Operating Characteristic curve; TPR = True Positive Rate; FPR = False Positive Rate. Source data are provided as a Source Data file.

Fig. 4. \textbf{Malignancy diagnosis performance of MOME on key subgroups on the combination of DS1 test set 1 and DS2.} AUROC, AUPRC, sensitivity, and specificity are reported for each subgroup, from left to right. Red and green bars at the right represent the number of malignant and benign cases for each subgroup. All metrics are presented with 95\% CI based on 1000-time bootstrap. AUROC = area under the Receiver Operating Characteristic curve; AUPRC = area under the precision-recall curve; BI-RADS = Breast Imaging-Reporting and Data System; BPE = Background Parenchymal Enhancement. DS1 = dataset 1; DS2 = dataset 2. Source data are provided as a Source Data file.

Fig. 5. \textbf{Decision Interpretation of MOME.} The illustrations correspond to DCE subtraction 3D visualization \revise{overlaid with saliencies in red computed by integrated gradient} (a,b,g,h), the zoomed-in axial view of DCE subtraction, DWI, and T2WI (c,d,i,j), the local Shapley value (e,f,k,l), and global Shapley value of the DS1 internal testing set (m) and DS2 (n). Four typical cases of a BI-RADS 5 patient with a malignant lesion (a,c,e), a BI-RADS 4 patient with a benign lesion (b,d,f) from DS1 internal testing set, and a BI-RADS 5 patient with a malignant lesion (g,i,k), a BI-RADS 4 patient with a benign lesion (h,j,i) from DS2 are shown. BI-RADS = Breast Imaging-Reporting and Data System; DCE = dynamic contrast-enhanced; DWI = Diffusion-weighted Imaging; T2WI = T2-weighted imaging. Source data are provided as a Source Data file.

Fig. 6. \textbf{Potential of noninvasive personalized treatment based on MOME.} The decision curves on DS1 internal testing set 1 (a), DS1 internal testing set 2 (b), DS2 (c), and DS3 (d) show a long range of preference in using MOME for malignancy screening. The decision curve also shows high net benefit of reducing biopsy for BI-RADS 4 patients on DS2 (e). MOME also demonstrated potential with ROC curves for TNBC patient subtyping (f) and NACT response prediction (g). Results are shown with 95\% CIs based on 1000-time bootstrap. Error bars represents the 95\% CIs. TPR = True Positive Rate; FPR = False Positive Rate; Rad = Radiologist. Source data are provided as a Source Data file.

\newpage

\begin{appendices}
\section*{Supplementary}\label{secA1}

\begin{figure}[ht]
    \centering
\includegraphics[width=\textwidth]{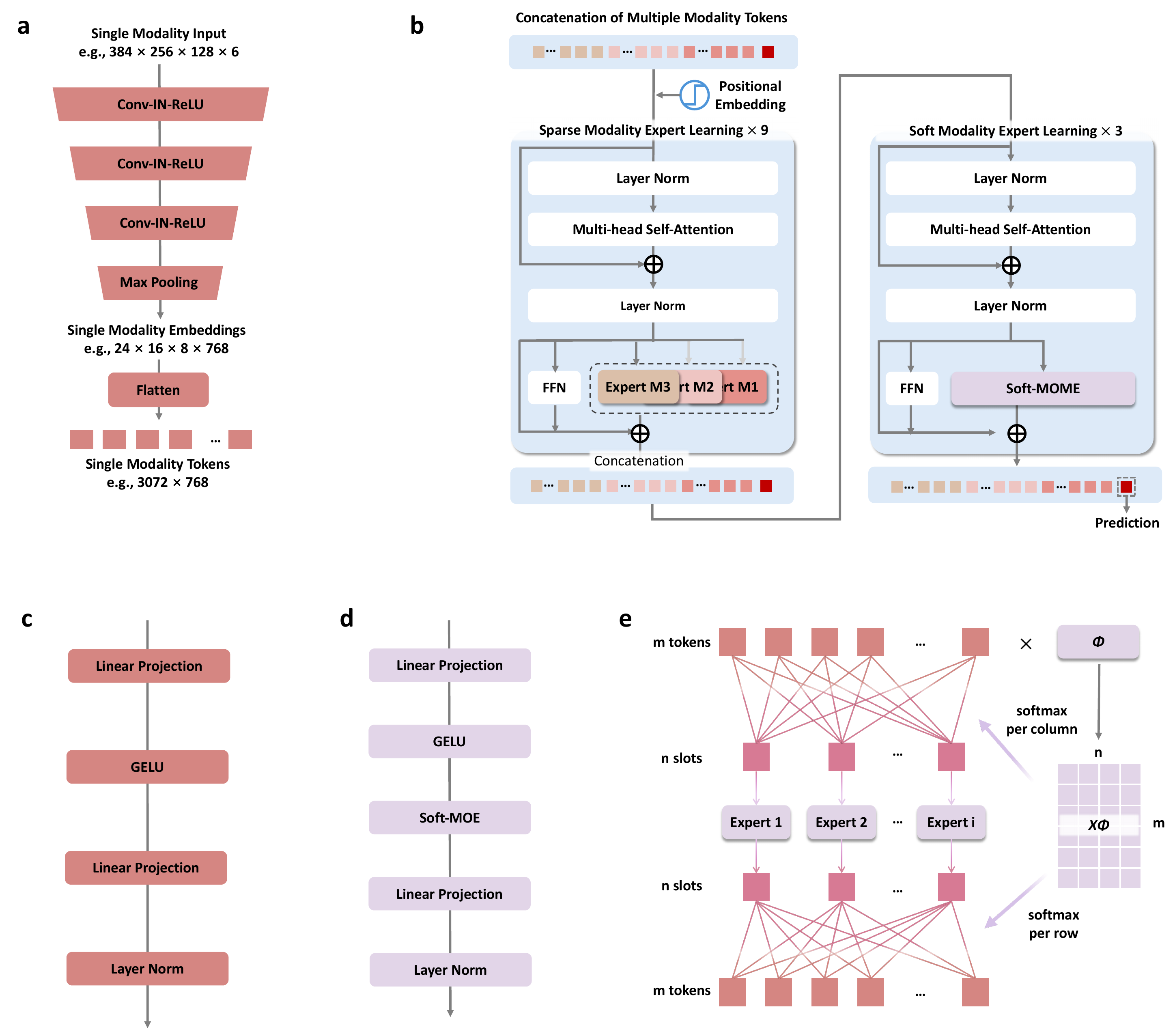}
    \caption{\revise{\textbf{Detailed structure of MOME}. a. The tokenizer structure, using one single modality branch as an example. There are in total three tokenizers for DCE-MRI, T2WI, and DWI, respectively. b. The structure of the transformer part of MOME, which was adapted from BEiT3 \cite{wang2023image}. All parameters are fixed except for the mixture of experts. c. The sparse MOME structure, using the DCE-MRI branch as an example. d. The soft MOME structure. e. The soft mixture of experts.}}
    \label{fig:dataflow}
\end{figure}

\newpage

\begin{figure}[ht]
    \centering
\includegraphics[width=\textwidth]{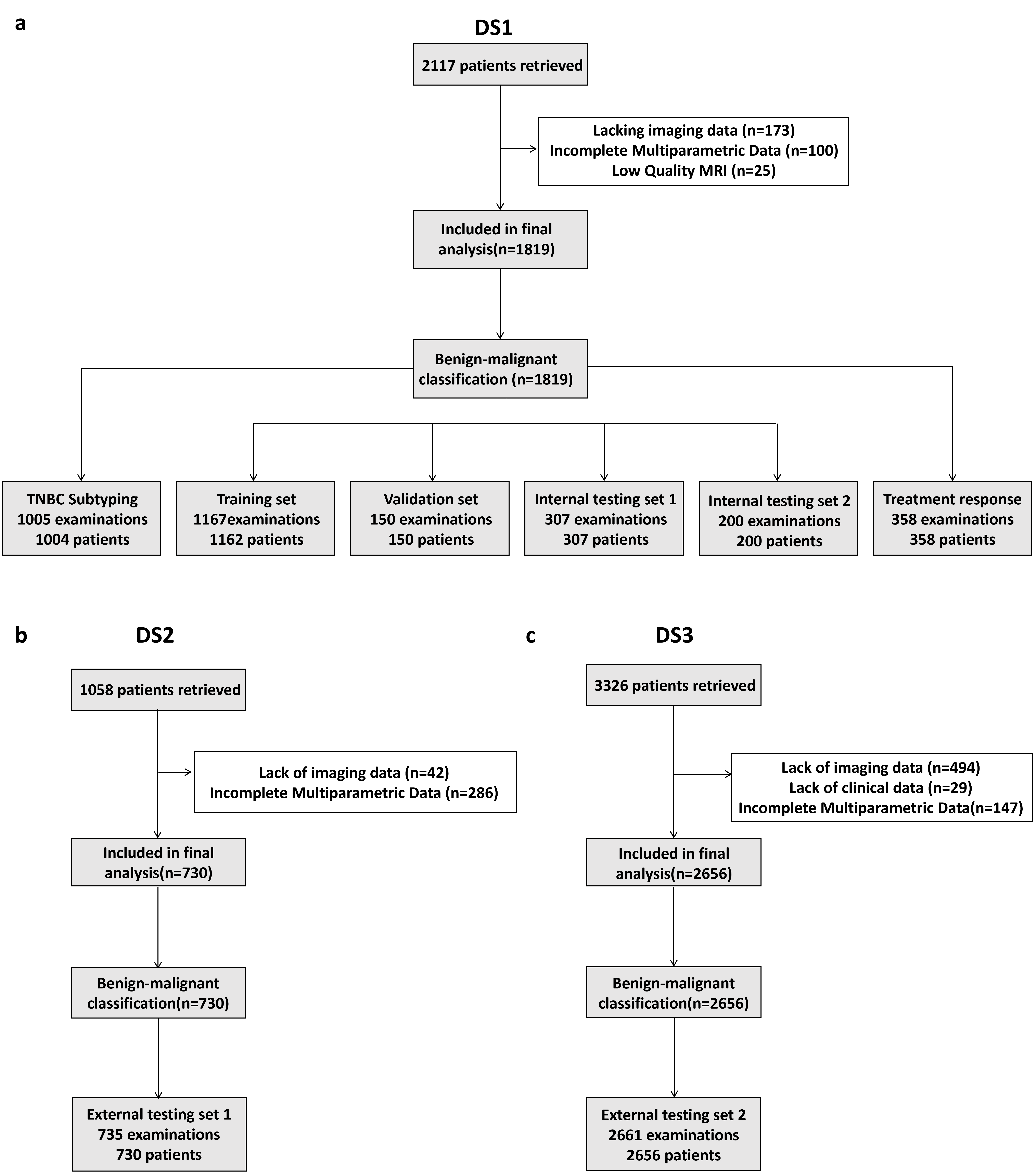}
    \caption{\textbf{Data collection flow chart for DS1 (a), DS2 (b), and DS3 (c)}.}
    \label{fig:dataflow}
\end{figure}

\newpage

\begin{table}[ht]
  \centering
  \label{tab:statistics} 
  \begin{tabular}{lcccc} 
  \toprule 
  
   \textbf{Task} & \textbf{Characteristics}& \textbf{DS1} & \textbf{DS2}& \textbf{DS3} \\
  \midrule 
  \textbf{Malignancy Classification}& & & & \\
   &\# Patients & 1,819 & 730& 2,656 \\
   &\# Examinations & 1,824 & 735& 2,661 \\
   &Age & 47.2$\pm$11.8& 47.3$\pm$10.5&46.4$\pm$11.8 \\
  & benign&638 &359&1259\\
   &malignant& 1186&376&1402\\
  & Unknown BI-RADS&0 &6 &141 \\
  & & - &(83.3\%, 16.7\%)&(94.3\%, 5.7\%)\\
   &BI-RADS0&0 &1 &0 \\
   & &-& (100\%, 0\%)&- \\
   &BI-RADS1&0 &30 &42 \\
    & & -&(100\%, 0\%)&(92.9\%, 7.1\%)\\
   &BI-RADS2&19 &55 &109 \\
    & & (100\%, 0\%)&(96.4\%, 3.6\%)&(97.2\%, 2.8\%)\\
   &BI-RADS3&370 &160 &515 \\
    & &(95.1\%, 4.9\%)&(98.8\%, 1.2\%)&(98.1\%, 1.9\%)\\
   &BI-RADS4&308 &195 &670 \\
    & &(67.9\%, 32.1\%)&(55.9\%, 44.1\%)&(67.9\%, 32.1\%)\\
   &BI-RADS5&1127 &53 &1,181 \\
    & &(5.1\%, 94.9\%)&(5.7\%, 94.3\%)&(1.8\%, 98.2\%)\\
   &BI-RADS6&0 &235 &3 \\
    & &-&(0\%, 100\%)&(0\%, 100\%)\\
  \textbf{Treatment Response}& & & & \\
   & \# Patients&358& & \\
   & \# Examinations&358& & \\
    &Age  
   & 47.3$\pm$10.5& & \\
   &pCR&90&-&-  \\
   &non-pCR&268&-&-  \\
  
  \textbf{Subtyping}& & & & \\
  & \# Patients&1,004& & \\
  & \# Examinations&1005& & \\
  
   &Age & 
   50.0$\pm$11.0& & \\
  & TNBC&60 &-&-  \\
   &non-TNBC&945 &-&-  \\
  \bottomrule 
  \end{tabular}
  \caption{\textbf{Patient characteristics in DS1, DS2 and DS3.} The numbers in the brackets represent the ratio of benign and malignant patients.} 
\end{table}

\newpage

\begin{table}[ht]
\centering
\label{tab:DS2} 
\begin{tabular}{cccccccccc} 
\toprule

 \textbf{Dataset}&\textbf{Scan Parameters} & \multicolumn{2}{c}{\textbf{NFS T1WI}} & \multicolumn{2}{c}{\textbf{T2WI}}& \multicolumn{2}{c}{\textbf{DWI}}& \multicolumn{2}{c}{\textbf{DCE-MRI}} \\
\midrule 
DS1& & & & & & & & & \\
 &TR (ms) & \multicolumn{2}{c}{8.7} & \multicolumn{2}{c}{2900} & \multicolumn{2}{c}{6200}& \multicolumn{2}{c}{4.53}\\
 &TE (ms) & \multicolumn{2}{c}{4.7} & \multicolumn{2}{c}{60} &\multicolumn{2}{c}{104}& \multicolumn{2}{c}{1.66}\\
 &slice thickness (mm) & \multicolumn{2}{c}{1} &\multicolumn{2}{c}{4} &\multicolumn{2}{c}{4}& \multicolumn{2}{c}{1}\\
 &matrix &\multicolumn{2}{c}{896$\times$896} & \multicolumn{2}{c}{640$\times$640}& \multicolumn{2}{c}{236$\times$120}& \multicolumn{2}{c}{384$\times$384}\\
 &b-value (s/mm$^2$) & \multicolumn{2}{c}{-}&\multicolumn{2}{c}{-}& \multicolumn{2}{c}{400/800/1000}&\multicolumn{2}{c}{-}\\
 &post-contrast scans & \multicolumn{2}{c}{-} & \multicolumn{2}{c}{-} & \multicolumn{2}{c}{-}& \multicolumn{2}{c}{6}\\
 \midrule 

DS2& & & & & & & & & \\
 & Scanner & 1.5T&3.0T&1.5T&3.0T&1.5T&3.0T&1.5T&3.0T\\

 &TR (ms) & 559 & 6 & 4500& 4740&6400&5700&5.2&4.7\\
 &TE (ms) & 12 & 2.5 & 102&107& 97&59&2.4&1.7\\
 &slice thickness (mm) & 4&1.6 & 4 & 4&4&4&1.1& 1.6\\
 &matrix & 448$\times$448 &448$\times$448 & 512$\times$512& 448$\times$448& 192$\times$192&
340$\times$170&
384$\times$384&
448$\times$448\\
 &b-value (s/mm$^2$) & -& -& -&-&50/500/1000&50/400/800& -&-\\
 &post-contrast scans & - & - & - &- &-&- & 5&5\\
\midrule

DS3& & & & & & & & & \\
 &TR (ms) & \multicolumn{2}{c}{8.6} & \multicolumn{2}{c}{5600} & \multicolumn{2}{c}{4900}& \multicolumn{2}{c}{4.43}\\
 &TE (ms) & \multicolumn{2}{c}{4.7} &\multicolumn{2}{c}{56} &\multicolumn{2}{c}{84}& \multicolumn{2}{c}{1.5}\\
 &slice thickness (mm) & \multicolumn{2}{c}{1} & \multicolumn{2}{c}{4} & \multicolumn{2}{c}{4}& \multicolumn{2}{c}{1.7}\\
 &matrix &\multicolumn{2}{c}{352$\times$324} & \multicolumn{2}{c}{320$\times$320}& 
  \multicolumn{2}{c}{220$\times$220}& \multicolumn{2}{c}{352$\times$324}\\
 &b-value (s/mm$^2$) & \multicolumn{2}{c}{-}& \multicolumn{2}{c}{-}&\multicolumn{2}{c}{0/800} & \multicolumn{2}{c}{-}\\
 &post-contrast scans & \multicolumn{2}{c}{-} & \multicolumn{2}{c}{-} & \multicolumn{2}{c}{-}& \multicolumn{2}{c}{5}\\

\bottomrule 
\end{tabular}
\caption{\textbf{Scanning parameters of MRI sequences.}} 
\end{table}

\newpage

\begin{table}[ht]
    \centering
    \resizebox{\textwidth}{!}{\begin{tabular}{cccccccc}
        \toprule
        \textbf{Radiologist} & \textbf{Accuracy} & \textbf{F1} & \textbf{MCC} & \textbf{Sensitivity} & \textbf{Specificity} & \textbf{PPV} & \textbf{NPV} \\
        \midrule
        \textbf{Radiologist 1} & 0.780 & 0.840 & 0.495 & 0.886 & 0.579 & 0.799 & 0.728 \\
        ($<$ 5 years) & (0.720, 0.835) & (0.789, 0.886) & (0.361, 0.615) & (0.833, 0.938) & (0.458, 0.688) & (0.725, 0.862) & (0.604, 0.843) \\
        \textbf{Radiologist 2} & 0.855 & 0.889 & 0.681 & 0.886 & 0.798 & 0.892 & 0.787 \\
        (5-10 years) & (0.800, 0.900) & (0.845, 0.925) & (0.569, 0.779) & (0.829, 0.937) & (0.695, 0.890) & (0.835, 0.940) & (0.689, 0.882) \\
        \textbf{Radiologist 3} & 0.830 & 0.861 & 0.651 & 0.809 & 0.868 & 0.921 & 0.706 \\
        (5-10 years) & (0.770, 0.880) & (0.808, 0.905) & (0.538, 0.757) & (0.737, 0.878) & (0.783, 0.940) & (0.870, 0.964) & (0.602, 0.809) \\
        \textbf{Radiologist 4} & 0.940 & 0.956 & 0.868 & 0.986 & 0.854 & 0.928 & 0.969 \\
        ($>$10 years) & (0.905, 0.970) & (0.927, 0.978) & (0.787, 0.933) & (0.962, 1.000) & (0.764, 0.933) & (0.881, 0.968) & (0.918, 1.000) \\
        \textbf{Radiologist 5} & 0.870 & 0.902 & 0.707 & 0.923 & 0.768 & 0.883 & 0.841 \\
        ($>$10 years) & (0.820, 0.915) & (0.864, 0.939) & (0.600, 0.808) & (0.874, 0.966) & (0.671, 0.862) & (0.828, 0.934) & (0.754, 0.923) \\
        \textbf{Radiologist 6} & 0.850 & 0.892 & 0.661 & 0.947 & 0.666 & 0.843 & 0.870 \\
        ($>$ 10 years) & (0.800, 0.895) & (0.853, 0.926) & (0.542, 0.757) & (0.902, 0.980) & (0.550, 0.776) & (0.781, 0.901) & (0.771, 0.950) \\
        \midrule
        \textbf{MOME} & 0.875 & 0.905 & 0.723 & 0.909 & 0.812 & 0.901 & 0.824 \\
        & (0.830, 0.920) & (0.868, 0.939) & (0.614, 0.818) & (0.857, 0.955) & (0.0.708, 0.899) & (0.849, 0.950) & (0.730, 0.910) \\
        \bottomrule
    \end{tabular}}
    \caption{\textbf{Detailed performance of radiologists and MOME on the DS1 internal testing set 2.} Radiologists are numbered and their experience in breast radiology is recorded in years. Results are reported as the mean and 95\% CI with 1000-time bootstrap.}
\end{table}

\newpage

\begin{table}[ht]
    \resizebox{\textwidth}{!}{
    \begin{tabular}{cccccccccc}
    \hline
    Model         & AUROC   & AUPRC & Accuracy & F1-score & Sensitivity & Specificity & PPV & NPV \\
    \hline
    \multirow{2}{*}{ResNet18-T2}      & 0.650 & 0.768 & 0.680 & 0.779 & 0.901 & 0.305 & 0.687 & 0.646 \\
    & 0.585, 0.709 & 0.705, 0.824 & 0.629, 0.733 & 0.735, 0.821 & 0.855, 0.942 & 0.221, 0.391 & 0.629, 0.744 & 0.516, 0.774 \\
    \multirow{2}{*}{ResNet18-DWI}      & 0.748 & 0.836 & 0.703 & 0.766 & 0.777 & 0.578 & 0.757 & 0.605 \\
    & 0.694, 0.803 & 0.782, 0.883 & 0.651, 0.752 & 0.718, 0.810 & 0.716, 0.835 & 0.496, 0.667 & 0.699, 0.814 & 0.517, 0.694 \\
    \multirow{2}{*}{ResNet34-DCE}      & 0.882 & 0.927 & 0.803 & 0.853 & 0.911 & 0.621 & 0.803 & 0.805 \\
    & 0.839, 0.918 & 0.891, 0.956 & 0.759, 0.844 & 0.815, 0.887 & 0.870, 0.947 & 0.539, 0.708 & 0.750, 0.852 & 0.721, 0.881 \\
    
    \multirow{2}{*}{LateFuse}      & 0.858 & 0.908 & 0.767 & 0.834 & {0.932} & 0.489 & 0.755 & {0.809} \\
    & 0.817, 0.897 & 0.868, 0.941 & 0.720, 0.811 & 0.794, 0.870 & 0.897, 0.965 & 0.404, 0.574 & 0.699, 0.807 & 0.719, 0.898 \\
    \multirow{2}{*}{Feature Fuse}  & 0.884 & 0.932 & 0.814 & 0.854 & 0.869 & 0.720 & 0.840 & 0.765 \\
    & 0.847, 0.918 & 0.896, 0.959 & 0.772, 0.857 & 0.817, 0.889 & 0.823, 0.914 & 0.639, 0.800 &  0.788, 0.889 & 0.683, 0.839 \\
    \multirow{2}{*}{BEiT3}         & 0.884 & 0.917 & 0.810 & 0.855 & 0.890 & 0.674 & 0.822 & 0.784 \\
    & 0.842, 0.922 & 0.869, 0.957 & 0.765, 0.853 & 0.816, 0.889 & 0.845, 0.929 & 0.591, 0.763 & 0.770, 0.874 & 0.699, 0.862 \\
    \multirow{2}{*}{MOME }     & \textbf{0.903} & \textbf{0.941} & {0.833} & {0.869} & 0.889 & {0.737} & {0.851} & 0.798 \\
    & 0.866, 0.936 & 0.910, 0.965 & 0.788, 0.870 & 0.831, 0.902 & 0.844, 0.929 & 0.655, 0.810 & 0.802, 0.899 & 0.717, 0.871 \\
    \hline

    \multirow{2}{*}{ResNet18-T2}  & 0.579 & 0.579 & 0.513 & 0.678 & {1.000} & 0.000 & 0.513 & - \\
    & 0.535, 0.617 & 0.525, 0.632 & 0.480, 0.547 & 0.649, 0.707 & 1.000, 1.000 & 0.000, 0.000 & 0.480, 0.547 & - \\
    \multirow{2}{*}{ResNet18-DWI}      & 0.442 & 0.467 & 0.514 & 0.677 & 0.995 & 0.008 & 0.514 & -\\
    & 0.402, 0.485 & 0.427, 0.512 & 0.480, 0.550 & 0.647, 0.707  & 0.986, 1.000 & 0.000, 0.019  & 0.479, 0.549 & -\\
    \multirow{2}{*}{ResNet34-DCE}      & 0.872 & 0.858 & {0.818} & {0.825} & 0.840 & 0.796 & 0.812 & 0.825 \\
    & 0.846, 0.898 & 0.816, 0.896 & 0.790, 0.846 & 0.795, 0.853  & 0.802, 0.875 & 0.752, 0.836 & 0.773, 0.849 & 0.787, 0.862  \\
    
    \multirow{2}{*}{LateFuse}     & 0.858 & 0.847 & 0.514 & 0.678 & {1.000} & 0.003 & 0.513 & -\\
    & 0.829, 0.885 & 0.805, 0.885 & 0.480, 0.548 & 0.649, 0.708 & 1.000, 1.000 & 0.000, 0.009 & 0.480, 0.548 & - \\
    \multirow{2}{*}{Feature Fuse} & 0.860 & 0.874 & 0.801 & 0.786 & 0.715 & {0.891} & {0.874} & 0.748  \\
    & 0.834, 0.886 & 0.840, 0.904  & 0.773, 0.829 & 0.751, 0.820 & 0.670, 0.760 & 0.859, 0.922 & 0.835, 0.909 & 0.708, 0.789  \\
    \multirow{2}{*}{BEiT3}         & 0.843 & 0.834 & 0.752 & 0.787 & 0.896 & 0.601 & 0.703 & {0.846} \\
    & 0.814, 0.871 & 0.793, 0.874 & 0.720, 0.782 & 0.757, 0.817 & 0.865, 0.924 & 0.548, 0.650 & 0.663, 0.743 & 0.799, 0.885  \\
    \multirow{2}{*}{MOME }     & \textbf{0.893} & \textbf{0.882} & 0.810 & 0.814  & 0.810 & 0.810 & 0.818 & 0.802 \\
    & 0.870, 0.916 & 0.840, 0.920  & 0.785, 0.838 & 0.785, 0.843 & 0.770, 0.854 & 0.770, 0.848  & 0.782, 0.856 & 0.761, 0.846  \\
    \hline

    \multirow{2}{*}{ResNet18-T2}  & \revise{0.549} & \revise{0.597} & \revise{0.561} & \revise{0.706} & \revise{0.967} & \revise{0.074} & \revise{0.557} & \revise{0.651} \\
& \revise{0.513}, \revise{0.581} & \revise{0.555}, \revise{0.639} & \revise{0.531}, \revise{0.591} & \revise{0.680}, \revise{0.731} & \revise{0.951}, \revise{0.980} & \revise{0.049}, \revise{0.099} & \revise{0.526}, \revise{0.587} & \revise{0.509}, \revise{0.781} \\
\multirow{2}{*}{ResNet18-DWI}      & \revise{0.478} & \revise{0.513} & \revise{0.569} & \revise{0.700} & \revise{0.921} & \revise{0.147} & \revise{0.565} & \revise{0.607} \\
& \revise{0.445}, \revise{0.514} & \revise{0.476}, \revise{0.552} & \revise{0.537}, \revise{0.599} & \revise{0.673}, \revise{0.726} & \revise{0.899}, \revise{0.941} & \revise{0.115}, \revise{0.178} & \revise{0.533}, \revise{0.596} & \revise{0.513}, \revise{0.695} \\
\multirow{2}{*}{ResNet34-DCE}      & \revise{0.873} & \revise{0.882} & \revise{0.815} & \revise{0.836} & \revise{0.865} & \revise{0.755} & \revise{0.809} & \revise{0.823} \\
& \revise{0.852}, \revise{0.895} & \revise{0.854}, \revise{0.909} & \revise{0.790}, \revise{0.838} & \revise{0.812}, \revise{0.859} & \revise{0.836}, \revise{0.892} & \revise{0.717}, \revise{0.791} & \revise{0.776}, \revise{0.841} & \revise{0.787}, \revise{0.859} \\
\multirow{2}{*}{LateFuse}     & \final{0.821} & \final{0.842} & \final{0.588} & \final{0.721} & \final{0.978} & \final{0.121} & \final{0.572} & \final{0.817} \\
& \final{0.794}, \final{0.847} & \final{0.809}, \final{0.872} & \final{0.559}, \final{0.619} & \final{0.695}, \final{0.747} & \final{0.965}, \final{0.989} & \final{0.092}, \final{0.150} & \final{0.540}, \final{0.603} & \final{0.726}, \final{0.908} \\
\multirow{2}{*}{Feature Fuse} & \revise{0.866} & \revise{0.893} & \revise{0.806} & \revise{0.812} & \revise{0.769} & \revise{0.850} & \revise{0.860} & \revise{0.754} \\
& \revise{0.843} \revise{0.888} & \revise{0.868} \revise{0.916} & \revise{0.780} \revise{0.830} & \revise{0.786} \revise{0.838} & \revise{0.732} \revise{0.806} & \revise{0.817} \revise{0.881} & \revise{0.830} \revise{0.889} & \revise{0.716} \revise{0.792} \\
\multirow{2}{*}{BEiT3}         & \revise{0.856} & \revise{0.863} & \revise{0.770} & \revise{0.809} & \revise{0.895} & \revise{0.620} & \revise{0.739} & \revise{0.831} \\
& \revise{0.833}, \revise{0.879} & \revise{0.830}, \revise{0.894} & \revise{0.745}, \revise{0.796} & \revise{0.784}, \revise{0.834} & \revise{0.869}, \revise{0.920} & \revise{0.576}, \revise{0.660} & \revise{0.705}, \revise{0.774} & \revise{0.793}, \revise{0.873} \\
\multirow{2}{*}{MOME}     & \textbf{\revise{0.896}} & \textbf{\revise{0.901}} & \revise{0.818} & \revise{0.834} & \revise{0.839} & \revise{0.793} & \revise{0.829} & \revise{0.804} \\
& \revise{0.876}, \revise{0.913} & \revise{0.873}, \revise{0.927} & \revise{0.794}, \revise{0.841} & \revise{0.809}, \revise{0.856} & \revise{0.807}, \revise{0.868} & \revise{0.753}, \revise{0.831} & \revise{0.798}, \revise{0.861} & \revise{0.767}, \revise{0.838} \\
\hline
\hline

    \end{tabular}
    }
\caption{\revise{\textbf{Detailed performance comparison between MOME and other approaches for breast cancer diagnosis.} The upper part of the table shows the results on DS1 internal testing set 1, the middle part of the table shows the results on DS2, and the lower part of the table shows the results on the combination of DS1 test set 1 and DS2. Results are reported as the mean and 95\% CI with 1000-time bootstrap. Best AUROC and best AUPRC are emphasized in bold. Results are obtained without test-time augmentation.}}
\label{tab:comparison_with_others}
\end{table}

\newpage

\begin{table}[ht]
  \centering
  \begin{tabular}{lcccc}
      \toprule
      \textbf{Metric} & \textbf{Internal Testing 1} & \textbf{Internal Testing 2} & \textbf{DS2} & \textbf{DS3} \\
      \midrule
      AUROC & 0.912  & 0.913  & 0.899  & 0.806  \\
            & (0.877, 0.944) & (0.864, 0.952) & (0.877, 0.922) & (0.790, 0.822) \\
      pAUROC (90\% specificity) & 0.800 & 0.765 & 0.753 & 0.621 \\
                                & (0.706, 0.880) & (0.650, 0.881) & (0.698, 0.805) & (0.600, 0.643) \\
      pAUROC (90\% sensitivity) & 0.735 & 0.752 & 0.740 & 0.617 \\
                                & (0.666, 0.811) & (0.642, 0.855) & (0.695, 0.788) & (0.594, 0.642) \\
      AUPRC & 0.942 & 0.948 & 0.887 & 0.807 \\
            & (0.907, 0.970) & (0.911, 0.977) & (0.847, 0.923) & (0.785, 0.827) \\
      Accuracy & 0.829 & 0.875 & 0.821 & 0.746 \\
            & (0.785, 0.873) & (0.830, 0.920) & (0.795, 0.849) & (0.729, 0.761) \\
      F1 & 0.866 & 0.905 & 0.828 & 0.763 \\
         & (0.828, 0.902) & (0.868, 0.939) & (0.798, 0.857) & (0.746, 0.779) \\
      Sensitivity & 0.884 & 0.909 & 0.839 & 0.780 \\
                  & (0.839, 0.929) & (0.857, 0.955) & (0.801, 0.877) & (0.759, 0.800) \\
      Specificity & 0.735 & 0.812 & 0.802 & 0.708 \\
                  & (0.655, 0.815) & (0.708, 0.899) & (0.762, 0.842) & (0.682, 0.732) \\
      PPV & 0.850 & 0.901 & 0.817 & 0.748 \\
          & (0.798, 0.898) & (0.849, 0.950) & (0.779, 0.854) & (0.725, 0.770) \\
      NPV & 0.790 & 0.824 & 0.826 & 0.743 \\
          & (0.712, 0.863) & (0.730, 0.910) & (0.787, 0.867) & (0.719, 0.766) \\
      \bottomrule
  \end{tabular}
  \caption{\textbf{Detailed performance of MOME on different datasets.} Results are reported as the mean and 95\% CI with 1000-time bootstrap.}
  \label{tab:performance_metrics}
\end{table}

\newpage

\begin{table}[ht]
    \begin{tabular}{cccccccc}
    \hline
    \textbf{Model} & \textbf{Inference} & \textbf{Accuracy} & \textbf{F1-score} & \textbf{Sensitivity} & \textbf{Specificity} & \textbf{PPV} & \textbf{NPV} \\
    \hline
    \multirow{2}{*}{w/o MOME} & \multirow{2}{*}{Multiparametric} & 0.823 & 0.860 & 0.870 & 0.743 & 0.851 & 0.771 \\
    & & 0.779, 0.866 & 0.821, 0.895 & 0.820, 0.914 & 0.664, 0.822 & 0.799, 0.901 & 0.686, 0.845 \\
    \multirow{2}{*}{w/o Soft MOME} & \multirow{2}{*}{Multiparametric} & 0.771 & 0.832 & {0.906} & 0.543 & 0.770 & 0.773 \\
    & & 0.726, 0.818 & 0.793, 0.869 & 0.866, 0.943 & 0.453, 0.637 & 0.716, 0.822 & 0.682, 0.863 \\
    \multirow{4}{*}{MOME} & \multirow{2}{*}{DCE} & 0.729 & 0.744 & 0.631 & {0.894} & {0.909} & 0.589 \\
    & & 0.678, 0.775 & 0.690, 0.794 & 0.566, 0.697 & 0.836, 0.944 & 0.857, 0.953 & 0.511, 0.659 \\
    & \multirow{2}{*}{Multiparametric} & {0.833} & {0.869} & 0.889 & 0.737 & 0.851 & 0.798 \\
    & & 0.788, 0.870 & 0.831, 0.902 & 0.844, 0.929 & 0.655, 0.810 & 0.802, 0.899 & 0.717, 0.871 \\
    \hline
    \end{tabular}
    \caption{\textbf{Other performance metrics in ablation study on MOME.} Results are reported as the mean and 95\% CI with 1000-time bootstrap. Best performance is emphasized in bold. Results are obtained without test-time augmentation.}
    \label{tab:ablation_other_metrics}
\end{table}

\newpage

\begin{table}[ht]
    \centering
    \begin{tabular}{lccccc}
    \toprule
    \textbf{Metric} & \textbf{Internal Testing 1} & \textbf{Internal Testing 2} & \textbf{DS2} & \textbf{DS3}\\
    \midrule
    AUROC & 0.886 & 0.897 & 0.881 & 0.790\\
        & (0.845, 0.920) & (0.839, 0.944) & (0.858, 0.906) & (0.772, 0.806)\\
    pAUROC (90\% specificity) & 0.752 & 0.717 & 0.726 & 0.617\\
            & (0.691, 0.809) & (0.558, 0.878) & (0.679, 0.775) & (0.597, 0.636)\\
    pAUROC (90\% sensitivity) & 0.691 & 0.718 & 0.702 & 0.569\\
            & (0.611, 0.778) & (0.604, 0.836) & (0.652, 0.755) & (0.545, 0.593)\\
    AUPRC & 0.933 & 0.926 & 0.882 & 0.803\\
                      & (0.903, 0.957) & (0.869, 0.972) & (0.849, 0.913) & (0.782, 0.822)\\
    Accuracy & 0.745 & 0.439 & 0.669 & 0.611\\
             & (0.694, 0.792) & (0.375, 0.505) & (0.635, 0.703) & (0.593, 0.630)\\
    F1 & 0.762 & 0.269 & 0.537 & 0.454\\
       & (0.705, 0.810) & (0.177, 0.365) & (0.487, 0.587) & (0.425, 0.481)\\
    Sensitivity & 0.652 & 0.158 & 0.375 & 0.306\\
        & (0.582, 0.716) & (0.098, 0.225) & (0.329, 0.425) & (0.282, 0.331)\\
    Specificity & 0.902 & 0.971 & 0.978 & 0.950\\
        & (0.846, 0.950) & (0.929, 1.000) & (0.962, 0.992) & (0.939, 0.962)\\
    PPV & 0.919 & 0.911 & 0.947 & 0.873\\
        & (0.867, 0.958) & (0.778, 1.000) & (0.910, 0.980) & (0.843, 0.901)\\
    NPV & 0.605 & 0.379 & 0.598 & 0.552\\
        & (0.527, 0.679) & (0.306, 0.448) & (0.559, 0.638) & (0.532, 0.573)\\
        \bottomrule
    \end{tabular}
\caption{\textbf{Detailed performance of MOME inferring only on DCE-MRI with test-time augmentation on different datasets.} Results are reported as the mean and 95\% CI with 1000-time bootstrap.}
\end{table}

\newpage

\begin{table}[ht]
    \centering
\label{tab:subgroup} 
    \begin{tabular}{lcccccc}
        \toprule
        \textbf{Subgroup} & \textbf{AUROC} & \textbf{AUPRC} & \textbf{Sensitivity} & \textbf{Specificity} & \textbf{\# Positives} & \textbf{\# Negatives} \\
        \midrule
        1.5T & 0.899 & 0.928 & 0.850 & 0.760 & 420 & 258 \\
        & (0.874, 0.921)    & (0.901, 0.952)    & (0.815, 0.884)    & (0.708, 0.811) & & \\
        3T & 0.897 & 0.873 & 0.882 & 0.799 & 144 & 139 \\
        & (0.856, 0.933)    & (0.800, 0.935)    & (0.832, 0.935)    & (0.730, 0.862) & & \\
        DS1 & 0.912 & 0.942 & 0.884 & 0.735 & 193 & 114 \\
        & (0.877, 0.944)    & (0.907, 0.970)    & (0.839, 0.929)    & (0.655, 0.815) & & \\
        DS2 & 0.899 & 0.887 & 0.839 & 0.802 & 376 & 359 \\
        & (0.877, 0.922)    & (0.847, 0.923)    & (0.801, 0.877)    & (0.762, 0.842) & & \\
        Age $<$ 40 & 0.875 & 0.807 & 0.838 & 0.731 & 92 & 165 \\
        & (0.832, 0.917)    & (0.714, 0.885)    & (0.760, 0.907)    & (0.661, 0.797) & & \\
        40 $\leq$ Age $<$ 60 & 0.902 & 0.900 & 0.843 & 0.815 & 364 & 291 \\
        & (0.878, 0.926)    & (0.859, 0.937)    & (0.807, 0.880)    & (0.769, 0.859) & & \\
        Age $\geq$ 60 & 0.938 & 0.989 & 0.912 & 0.825 & 113 & 17 \\
        & (0.872, 0.986)    & (0.975, 0.998)    & (0.857, 0.956)    & (0.615, 1.0) & & \\
        Minimal or Mild BPE & 0.906 & 0.922 & 0.866 & 0.782 & 438 & 322 \\
        & (0.885, 0.925)    & (0.896, 0.946)    & (0.836, 0.895)    & (0.738, 0.826) & & \\
        Moderate BPE & 0.894 & 0.859 & 0.833 & 0.828 & 96 & 99 \\
        & (0.840, 0.939)    & (0.767, 0.945)    & (0.750, 0.903)    & (0.748, 0.898) & & \\
        Marked BPE & 0.875 & 0.828 & 0.801 & 0.729 & 35 & 52 \\
        & (0.799, 0.937)    & (0.702, 0.918)    & (0.659, 0.917)    & (0.604, 0.846) & & \\
        BI-RADS 4 & 0.793 & 0.712 & 0.749 & 0.723 & 103 & 148 \\
        & (0.738, 0.847)    & (0.613, 0.811)    & (0.660, 0.828)    & (0.649, 0.797) & 	&  \\
        Other BI-RADS & 0.932 & 0.950 & 0.879 & 0.812 & 465 & 320 \\
        & (0.913, 0.949)    & (0.927, 0.968)    & (0.850, 0.908)    & (0.765, 0.853) & 	&  \\

        \bottomrule
    \end{tabular}
    \caption{\textbf{Detailed subgroup performance on the combination of DS1 test set 1 and DS2.} Results are reported as the mean and 95\% CI with 1000-time bootstrap.} 
\end{table}

\end{appendices}

\end{document}